\documentclass{article}

\usepackage[final]{corl_2020_plain} 

\title{Learning Equality Constraints for \\ Motion Planning on Manifolds}

\author{
Giovanni Sutanto*, 
Isabel M. Rayas Fern{\'a}ndez,
Peter Englert,\\
\textbf{Ragesh K. Ramachandran,
Gaurav S. Sukhatme}\\
Robotic Embedded Systems Laboratory\\ University of Southern California
}
\usepackage{graphics} 
\usepackage{epsfig} 
\usepackage{times} 
\usepackage{amsmath} 
\usepackage{amssymb}  
\usepackage{algorithm}
\usepackage[noend]{algpseudocode}
\usepackage{caption}
\usepackage{subcaption}

\usepackage[shortlabels]{enumitem}
\usepackage{multirow}
\usepackage{booktabs}
\usepackage[nocomma]{optidef}
\usepackage{cancel}
\usepackage{color}
\usepackage[normalem]{ulem}
\usepackage{comment}
\usepackage{sidecap}
\usepackage{wrapfig}
\usepackage{siunitx}

\DeclarePairedDelimiter\abs{\lvert}{\rvert}
\DeclarePairedDelimiter\norm{\lVert}{\rVert}

\makeatletter
\let\oldabs\abs
\def\abs{\@ifstar{\oldabs}{\oldabs*}}
\let\oldnorm\norm
\def\norm{\@ifstar{\oldnorm}{\oldnorm*}}
\makeatother

\newcommand{\ecmnnlong}{Equality Constraint Manifold Neural Network}
\newcommand{\ecmnn}{ECoMaNN}
\newcommand{\configspace}{\mathcal{C}}
\newcommand{\numdatapoints}{N}
\newcommand{\dimambient}{d}
\newcommand{\dimconstraint}{l}

\newcommand{\jacobian}{\mathbf{J}}
\newcommand{\jointposition}{\mathbf{q}}

\newcommand{\eye}{\mathbf{I}}
\newcommand{\numconstraintmanifold}{m}
\newcommand{\constraintmanifold}{M}
\newcommand{\constraintmanifoldsequence}{\mathcal{\constraintmanifold}}
\newcommand{\constraintfunction}{{h}_{\constraintmanifold}}
\newcommand{\constraintmanifoldjacobian}{\jacobian_\constraintmanifold}
\newcommand{\onconstraintconfigspace}{{\configspace}_{\constraintmanifold}}
\newcommand{\offconstraintconfigspace}{{\configspace}_{\backslash {\constraintmanifold}}}
\newcommand{\numnearestneighbor}{K}
\newcommand{\idxnearestneighbor}{k}
\newcommand{\KNNset}{\mathcal{K}}
\newcommand{\origKNN}{\hat{\KNNset}}
\newcommand{\recenteredKNN}{\tilde{\KNNset}}
\newcommand{\nearestneighborjointposition}{\hat{\jointposition}}
\newcommand{\recenterednearestneighborjointposition}{\tilde{\jointposition}}
\newcommand{\coordframe}{\mathcal{F}}
\newcommand{\orthobasisvec}{\vct{b}}
\newcommand{\orthobasismat}{\mathbf{B}}
\newcommand{\flippedorthobasismat}{\bar{\orthobasismat}}
\newcommand{\randunitvec}{\vct{u}}
\newcommand{\randscalarweight}{w}
\newcommand{\designmatrix}{\mathbf{X}}
\newcommand{\samplecovariancematrix}{\mathbf{S}}

\newcommand{\diagsingularvalues}{\mathbf{\Sigma}}
\newcommand{\eigval}{\lambda}
\newcommand{\righteigmat}{\mathbf{V}}
\newcommand{\covdiagsingularvalues}{\diagsingularvalues}
\newcommand{\coveigval}{\eigval}

\newcommand{\covrighteigmat}{\righteigmat}
\newcommand{\lpcacoordframe}{\coordframe_L}
\newcommand{\tangentspaceid}{T}
\newcommand{\normalspaceid}{N}
\newcommand{\tangentspaceatjointposition}{\tangentspaceid_{\jointposition}\constraintmanifold}
\newcommand{\normalspaceatjointposition}{\normalspaceid_{\jointposition}\constraintmanifold}
\newcommand{\lnormalcoordframe}{\coordframe_\normalspaceid}
\newcommand{\coveigvec}{\vct{v}}
\newcommand{\constrjaceigvec}{\vct{e}}
\newcommand{\nullspaceid}{\text{N}}
\newcommand{\covnullspaceeigmat}{\righteigmat_{\nullspaceid}}
\newcommand{\covnullspaceolmat}{\mathbf{W}_{\nullspaceid}}
\newcommand{\flippedrighteigmat}{\bar{\righteigmat}}
\newcommand{\flippedcovnullspaceeigmat}{\flippedrighteigmat_{\nullspaceid}}
\newcommand{\constrjacrighteigmat}{\mathbf{E}}
\newcommand{\constrjacnullspaceeigmat}{\constrjacrighteigmat_{\nullspaceid}}
\newcommand{\augjointposition}{\check{\jointposition}}
\newcommand{\augidx}{i}
\newcommand{\augmagnitude}{\epsilon}
\newcommand{\cost}{J}
\newcommand{\tpath}{\tau}
\newcommand{\tpathset}{\vct{\tpath}}

\newcommand{\tpathspace}{\mathcal{T}}
\newcommand{\freeconfigspaceoperator}{\Upsilon}
\newcommand{\loss}{\mathcal{L}}
\newcommand{\normloss}{\loss_\text{norm}}
\newcommand{\siamreflectionloss}{\loss_\text{reflection}}
\newcommand{\siamfracloss}{\loss_\text{fraction}}
\newcommand{\siamsimilarloss}{\loss_\text{similar}}
\newcommand{\osaloss}{\loss_\text{osa}}
\newcommand{\rotmat}{\mathbf{R}}
\newcommand{\diffson}{\rotmat_{\normalspaceid}}
\newcommand{\globalalignmentrotmat}{\rotmat_{G}}
\newcommand{\skewsymmmat}{\mathbf{L}}
\newcommand{\mstnumnearestneighbor}{H}
\newcommand{\nnsparsegraph}{\mathcal{G}}
\newcommand{\mst}{\mathcal{T}}
\newcommand{\dagedges}{\mathcal{E}}
\newcommand{\rootid}{r}
\newcommand{\rootjointposition}{\jointposition_{\rootid}}
\newcommand{\directededge}{\vct{e}}
\newcommand{\alignedcovnullspaceeigmat}{\covnullspaceeigmat^{aligned}}
\newcommand{\currentnodeid}{d}
\newcommand{\parentnodeid}{p}

\newcommand{\argmin}{\operatorname*{arg\:min}}
\newcommand{\argmax}{\operatorname*{arg\:max}}

\renewcommand{\Re}{\mathbb{R}}

\newcommand{\vct}[1]{\boldsymbol{#1}}

\newcommand{\T}{^{\textrm T}}

\newcommand{\period}{~.}
\renewcommand{\t}[1]{{\textrm{#1}}}

\newcommand{\st}{\t{s.t.}}
\newcommand{\lfd}{LfD}

\newcommand{\gsutast}{\bgroup\markoverwith{\textcolor{green}{\rule[0.5ex]{2pt}{0.4pt}}}\ULon}
\newcommand{\ragst}{\bgroup\markoverwith{\textcolor{magenta}{\rule[0.5ex]{2pt}{0.4pt}}}\ULon}
\newcommand{\new}[1]{#1}

\newcommand\blfootnote[1]{%
  \begingroup
  \renewcommand\thefootnote{}\footnote{#1}%
  \addtocounter{footnote}{-1}%
  \endgroup
}

\begin{document}
\maketitle

\begin{abstract}
\blfootnote{
\hspace{-1.7em}
* Giovanni Sutanto is now at X Development LLC. He contributed to this work during his past affiliation with the Robotic Embedded Systems Laboratory at USC.}
Constrained robot motion planning is a widely used technique to solve complex robot tasks. 
We consider the problem of learning representations of constraints from demonstrations with a deep neural network, which we call {\ecmnnlong} ({\ecmnn}). The key idea is to learn a level-set function of the constraint suitable for integration into a constrained sampling-based motion planner. Learning proceeds by aligning subspaces in the network with subspaces of the data.
We combine both learned constraints and analytically described constraints into the planner and use a projection-based strategy to find valid points. 
We evaluate {\ecmnn} on its representation capabilities of constraint manifolds, the impact of its individual loss terms, and the motions produced when incorporated into a planner.\\
\textbf{Video: } {\small{\url{https://www.youtube.com/watch?v=WoC7nqp4XNk}}}\\
\textbf{Code: } {\small{\url{https://github.com/gsutanto/smp_manifold_learning}}}
\end{abstract}


\section{Introduction}
\label{sec:introduction}

Robots must be able to plan motions that follow various constraints in order for them to be useful in real-world environments.
Constraints such as holding an object, maintaining an orientation, or staying within a certain distance of an object of interest are just some examples of possible restrictions on a robot's motion.
In general, two approaches to many robotics problems can be described. One is the traditional approach of using handwritten models to
capture environments, physics, and other aspects of the problem mathematically or analytically, and then solving or optimizing these to find a solution. 
The other, popularized more recently, involves the use of machine learning to replace, enhance, or simplify these hand-built parts.
Both have challenges: Acquiring training data for learning can be difficult and expensive, while describing precise models analytically can range from tedious to impossible.
Here, we approach the problem from a machine learning perspective and propose a solution to learn constraints from demonstrations. The learned constraints can be used alongside analytical solutions within a motion planning framework.

In this work, we propose a new learning-based method for describing motion constraints, called {\ecmnnlong} (\ecmnn). {\ecmnn} learns a function which evaluates a robot configuration on whether or not it meets the constraint, and for configurations near the constraint, on how far away it is. 
We train {\ecmnn} with datasets consisting of configurations that adhere to constraints, and present results for kinematic robot tasks learned from demonstrations. 
We use a sequential motion planning framework 
to solve motion planning problems that are both constrained and sequential in nature, 
and incorporate the learned constraint representations into it. We evaluate the constraints learned by \mbox{\ecmnn} with various datasets on their representation quality. Further, we investigate the usability of learned constraints in sequential motion planning problems.

\section{Related work}
\label{sec:related_work}

\subsection{Manifold learning}
Manifold learning is applicable to many fields and thus there exist a wide variety of methods for it. Linear methods include PCA and LDA \citep{duda2001pattern}, and while they are simple, they lack the complexity to represent complex manifolds. Nonlinear methods include multidimensional scaling (MDS),
locally linear embedding (LLE),
Isomap,
 and local tangent space alignment (LTSA).
These approaches use techniques such as eigenvalue decomposition, nearest neighbor reconstructions, and local-structure-preserving graphs to visualize and represent manifolds.
In LTSA, the local tangent space information of each point is aligned to create a global representation of the manifold. 
We refer the reader to \cite{ma2011manifold} for details.
Recent work in manifold learning additionally takes advantage of (deep) neural architectures. 
Some approaches use autoencoder-like models \citep{holden2015learning, chen2016dynamic} or deep neural networks \citep{nguyen2019neural} to learn manifolds, e.g. of human motion. 
Others use classical methods combined with neural networks, for example as a loss function for control \citep{sutanto2019learning} or as structure for the network \citep{wang2014generalized}.
Locally Smooth Manifold Learning (LSML) \citep{dollar2007non} defines and learns a function which describes the tangent space of the manifold, allowing randomly sampled points to be projected onto it. 
Our work is related to many of these approaches; in particular, the tangent space alignment in LTSA is an idea that {\ecmnn} uses extensively.
Similar to the ideas presented in this paper, the work in \cite{osa2020learning}  delineates an approach to solve motion planning problems by learning the solution manifold of an optimization problem. 
In contrast to others, our work focuses on learning implicit functions of equality constraint manifolds, which is a generalization of the learning representations for Signed Distance Fields (SDF) \cite{park2019deepsdf, mahler2015gp}, up to a scale, for higher-dimensional manifolds.

\subsection{Learning from demonstration}
Learning from demonstration (LfD) techniques learn a task representation from data which is usable to generate robot motions that imitate the demonstrated behavior.
One approach to LfD is inverse optimal control (IOC), which aims to find a cost function that describes the demonstrated behavior \cite{RatliffBZ06,ziebart2008maximum, ratliff2009learch, levine2012cioc}. 
Recently, IOC has been extended to extract constraints from demonstrations \cite{puydupin2012convex, englert2017ijrr}. 
 There, a cost function as well as equality and inequality constraints are extracted from demonstrations, which are useful to describe behavior like contacts or collision avoidance.
Our work can be seen as a special case where the task is only represented in form of constraints. Instead of using the extracted constraints in optimal control methods, we integrate them into 
sampling-based motion planning methods, which are not parameterized by time and do not suffer from poor initializations.
A more direct approach to LfD is to learn parameterized motion representations \cite{schaal2003computational,Paraschos2013pmp,Pastor_RAIIC_2011}. They represent the demonstrations in a parameterized form such as Dynamic Movement Primitives \cite{Ijspeert_NC_2013}. Here, learning a primitive from demonstration is often possible via linear regression; however, the ability to generalize to new situations is more limited. 
Other approaches to LfD include task space learning \cite{14-jetchev-AuRo} and deep learning \cite{finn2016guided}. We refer the reader to the survey \cite{argall2009survey} for a broad overview on LfD.

\subsection{Constrained sampling-based motion planning}
Sampling-based motion planning (SBMP) is a broad field which tackles the problem of motion planning by using randomized sampling techniques to build a tree or graph of configurations (also called samples), which can then be used to plan paths between configurations. 
Many SBMP algorithms derive from rapidly-exploring random trees (RRT) \citep{lavalle1998rapidly},  probabilistic roadmaps (PRM)
\citep{kavraki1994randomized}, or their optimal counterparts 
\citep{karaman2011sampling}. 
A more challenging and realistic motion planning task is that of constrained
SBMP \cite{kingston2018sampling}, where there are motion constraints beyond just obstacle avoidance which lead to a free configuration space manifold of lower dimension than the ambient configuration space.
Previous research has also investigated incorporating learned constraints or manifolds into planning frameworks. 
These include performing planning in learned latent spaces \cite{ichter2019robot}, learning a better sampling distribution in order to take advantage of the structure of valid configurations rather than blindly sample uniformly in the search space \citep{ichter2018learning, madaanlearning}, and  attempting to approximate the manifold (both explicitly and implicitly) of valid points with graphs in order to plan on them more effectively \citep{phillips2012graphs, havoutis2009motion, zha2018learning}. 
Our method differs from previous work in that {\ecmnn} learns an implicit description of a constraint manifold via a level set function, and during planning, we assume this representation for each task. We note that  our method could be combined with others, e.g. learned sampling distributions, to further improve planning results.

\section{Background}
\label{sec:background}

\subsection{Manifold theory}
\label{subsec:manifold_theory}

Here we present the necessary background on manifold theory \cite{Boothby:107707}. 
Informally, a manifold is a surface which can be well-approximated locally using an open set of a Euclidean space near every point.
Manifolds are generally defined using \textit{charts}, which are collections of open sets whose union yields the manifold, and a \textit{coordinate map}, which is a continuous map associated with each set. 
However, an alternative representation which is useful from a computational perspective is to represent the manifold as the zero level set of a continuous function. 
Since the latter representation is a direct result of the implicit function theorem, it is referred to as \textit{implicit representation} of the manifold. 
\new{For example, the manifold represented by the zero level set of the function $\constraintfunction(x,y) = x^2 + y^2-1$ (i.e. $\{ (x,y)~|~ x^2 + y^2-1 = 0\}$) is a circle.}
Moreover, the implicit function associated with a smooth manifold is smooth. 
\new{Thus, we can associate a manifold with every equality constraint.}
The vector space containing the set of all tangent vectors at $\jointposition$ is denoted using $\tangentspaceatjointposition$. Given a manifold with the corresponding implicit function $\constraintfunction(\jointposition)$, 
if we endow its tangent spaces with an appropriate inner product structure, then 
such a manifold is often referred as a Riemannian manifold. In this work the manifolds are assumed to be Riemannian. 

\subsection{Motion planning on manifolds}
\label{subsec:smp}
In this work, we aim to integrate learned constraint manifolds into 
a motion planning framework \citep{englert2020sampling}.
The motion planner considers kinematic motion planning problems in a configuration space $\configspace \subseteq \Re^{\dimambient}$. A robot configuration 
$\jointposition \in \configspace$
describes the state of one or more robots with 
$\dimambient$
degrees of freedom in total.
A manifold 
$\constraintmanifold$
is represented as an equality constraint 
$\constraintfunction(\jointposition) = \vct{0}$.
The set of robot configurations that are on a manifold 
$\constraintmanifold$
is given by 
\mbox{$\onconstraintconfigspace = \{ \jointposition \in \configspace \mid \constraintfunction(\jointposition) = \vct{0}\}\period$}
The planning problem is defined as a sequence of
$(\numconstraintmanifold+1)$
such manifolds 
\mbox{$\mathcal{\constraintmanifold} = \{\constraintmanifold_1, \constraintmanifold_2, \dots, \constraintmanifold_{\numconstraintmanifold+1}\}$}
and an initial configuration 
$\jointposition_\t{start} \in \configspace_{{\constraintmanifold}_1}$ 
on the first manifold.
The goal is to find a path from 
$\jointposition_\t{start}$ 
that traverses the manifold sequence 
$\constraintmanifoldsequence$
and reaches a configuration on the goal manifold 
$\constraintmanifold_{\numconstraintmanifold+1}$. 
A path on the $i$-th manifold is defined as 
$\tpath_i : [0, 1] \to 
\configspace_{{\constraintmanifold}_i}
$
and 
$\cost(\tpath_i)$ 
is the cost function of a path
$\cost : \tpathspace \to \Re_{\geq 0}$
where 
$\tpathspace$ 
is the set of all non-trivial paths. The problem is formulated as an optimization over a set of paths 
$\tpathset = (\tpath_1, \dots, \tpath_{\numconstraintmanifold})$ 
that minimizes the sum of path costs under the constraints of traversing 
$\constraintmanifoldsequence$:
\begin{align} 
	\begin{alignedat}{2} 
	\label{eq:smp_problem}
	&\qquad\qquad\qquad\qquad\qquad\tpathset^{\star} = \argmin_{\tpathset} \sum_{i=1}^{\numconstraintmanifold} \cost(\tpath_i) \\
	\st\quad &{\tpath}_1 (0) = {\jointposition}_\t{start}, \quad \tpath_{\numconstraintmanifold}(1) \in 
	\configspace_{{\constraintmanifold}_{\numconstraintmanifold+1}}
	\cap \configspace_{\t{free}, \numconstraintmanifold+1},\quad \tpath_i(1) = \tpath_{i+1}(0)\quad  \forall_{i=1,\dots,\numconstraintmanifold-1} \\ 
	&\configspace_{\t{free}, i+1} = \freeconfigspaceoperator(\configspace_{\t{free}, i}, \tpath_{i})\quad \forall_{i=1,\dots,\numconstraintmanifold},\quad
	\tpath_i(s) \in 
	\configspace_{{\constraintmanifold}_i}
	\cap \configspace_{\t{free}, i} \quad\forall_{i=1,\dots,\numconstraintmanifold}~ \forall_{s \in [0, 1]}
	\end{alignedat} 
\end{align}
$\freeconfigspaceoperator$
is an operator that describes the change in the free configuration space (the space of all configurations that are not in collision with the environment) 
$\configspace_\t{free}$ 
when transitioning to the next manifold. \new{The operator $\freeconfigspaceoperator$ is not explicitly known and we only assume to have access to a collision checker that depends on the current robot configuration and the object locations in the environment. Intelligently performing goal-achieving manipulations that change the free configuration space forms a key challenge in robot manipulation planning.}
\new{The SMP$^*$ (Sequential Manifold Planning) algorithm is able to solve this problem for a certain class of motion planning scenarios.} It iteratively applies RRT$^*$ to find a path that reaches the next manifold while staying on the current manifold.
For further details of the SMP$^*$ algorithm, we refer the reader to \citep{englert2020sampling}. 
In this paper, we employ data-driven learning methods to learn individual equality constraints \new{$\constraintfunction(\jointposition)=0$} 
from demonstrations with the goal to integrate them with analytically defined manifolds into 
this framework.

\section{{\ecmnnlong} (\ecmnn)}
\label{ssec:ecmnn}

We propose a novel neural network structure, called \emph{\ecmnnlong} (\ecmnn), which is a (global) equality constraint manifold learning representation that enforces the alignment of the (local) tangent spaces and normal spaces with the information extracted from the Local Principal Component Analysis (Local PCA) \citep{Kambhatla_LocalPCA} of the data. 
{\ecmnn} takes a configuration $\jointposition$ as input and outputs the prediction of the implicit function $\constraintfunction(\jointposition)$.
We train {\ecmnn} in a supervised manner, from demonstrations. 
One of the challenges is that the supervised training dataset is collected only from demonstrations of data points which are on the manifold $\onconstraintconfigspace$, called the \emph{on-manifold} dataset.
Collecting both the on-manifold $\onconstraintconfigspace$ and off-manifold $\offconstraintconfigspace = \{ \jointposition \in \configspace \mid \constraintfunction(\jointposition) \neq \vct{0} \}$ datasets for supervised training would be tedious because the implicit function $\constraintfunction$ values of points in $\offconstraintconfigspace$ are typically unknown and hard to label.
We show that, though our approach is only given data on $\onconstraintconfigspace$, it can still learn a useful and sufficient representation of the manifold for use in planning.

\begin{wrapfigure}{r}{0.25\textwidth}
    \centering
     \vspace{-10pt}
    \includegraphics[width=0.25\textwidth]{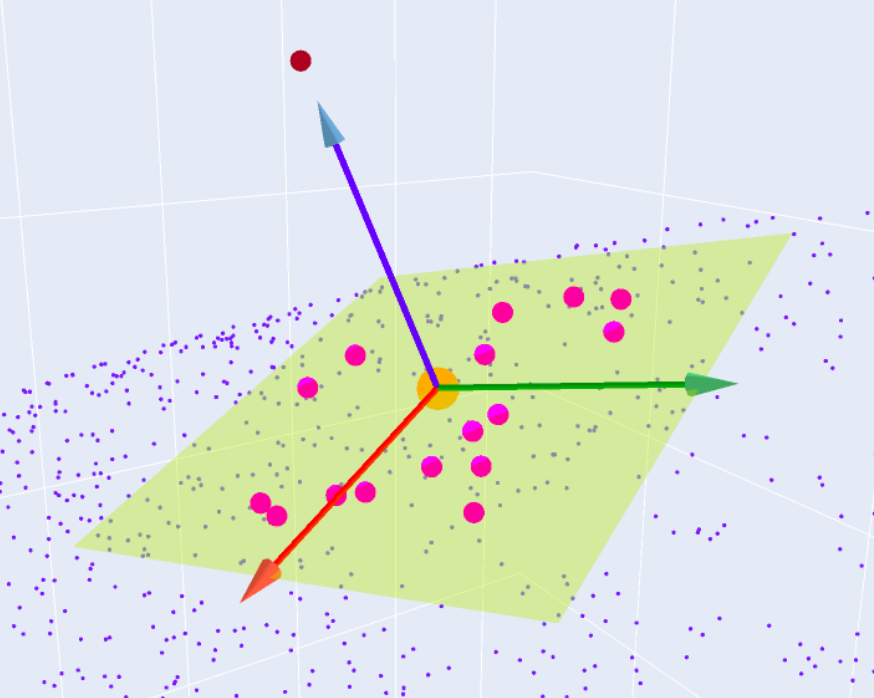}
    \caption{A visualization of data augmentation along the 1D normal space of a point $\jointposition$ in 3D space. Here, purple points are the dataset, pink points are the $\numnearestneighbor$NN of $\jointposition$, and the dark red point is $\augjointposition$. $\jointposition$ is at the axes origin, and the green plane is the approximated tangent space at that point.}
    \vspace{-25pt}
    \label{fig:localpca}
\end{wrapfigure}
Our goal is to learn a single global representation of the constraint manifold $M$ in the form of a neural network. 
Our approach leverages local information on the manifold in the form of the tangent and normal spaces \citep{Deutsch2015_TensorVotingGraph}.
With regard to $\constraintfunction$, the tangent and normal spaces are equivalent to the null and row space, respectively, of the Jacobian matrix 
$\constraintmanifoldjacobian(\acute{\jointposition}) = \left. \frac{\partial \constraintfunction({\jointposition})}{\partial {\jointposition}}\right|_{{\jointposition} = \acute{\jointposition}}$, 
and valid in a small neighborhood around the point $\acute{\jointposition}$.
Using on-manifold data, the local information of the manifold can be analyzed using Local PCA. For each data point $\jointposition$ in the on-manifold dataset, we establish a local neighborhood using $\numnearestneighbor$-nearest neighbors ($\numnearestneighbor$NN) $\origKNN = \{\nearestneighborjointposition_1, \nearestneighborjointposition_2, \dots, \nearestneighborjointposition_\numnearestneighbor\}$, with $\numnearestneighbor \geq \dimambient$. 
After a change of coordinates, $\jointposition$ becomes the origin of a new local coordinate frame $\lpcacoordframe$, and the $\numnearestneighbor$NN becomes $\recenteredKNN = \{\recenterednearestneighborjointposition_1, \recenterednearestneighborjointposition_2, \dots, \recenterednearestneighborjointposition_\numnearestneighbor\}$ with $\recenterednearestneighborjointposition_\idxnearestneighbor = \nearestneighborjointposition_\idxnearestneighbor - \jointposition$ for all values of $\idxnearestneighbor$. Defining the matrix 
$\designmatrix = 
\begin{bmatrix} 
\recenterednearestneighborjointposition_1 & \recenterednearestneighborjointposition_2 & \hdots & \recenterednearestneighborjointposition_\numnearestneighbor \\
\end{bmatrix}\T \in \Re^{\numnearestneighbor \times \dimambient}
$, we can compute the covariance matrix $\samplecovariancematrix = \frac{1}{\numnearestneighbor-1} \designmatrix\T \designmatrix \in \Re^{\dimambient \times \dimambient}$.
The eigendecomposition of $\samplecovariancematrix = \covrighteigmat \covdiagsingularvalues \covrighteigmat\T$ gives us the Local PCA. 
The matrix $\covrighteigmat$ contains the eigenvectors of $\samplecovariancematrix$ as its columns in decreasing order w.r.t.\ the corresponding eigenvalues in the diagonal matrix $\covdiagsingularvalues$. These eigenvectors form the basis of the coordinate frame $\lpcacoordframe$.

This local coordinate frame $\lpcacoordframe$ is tightly related to the tangent space $\tangentspaceatjointposition$ and the normal space $\normalspaceatjointposition$ of the manifold $\constraintmanifold$ at $\jointposition$. 
That is, the $(\dimambient - \dimconstraint)$ eigenvectors corresponding to the $(\dimambient - \dimconstraint)$ biggest eigenvalues of $\covdiagsingularvalues$ form a basis of $\tangentspaceatjointposition$, while the remaining $\dimconstraint$ eigenvectors form the basis of $\normalspaceatjointposition$. 
Furthermore, 
the $\dimconstraint$ smallest eigenvalues of $\covdiagsingularvalues$ will be close to zero, resulting in the $\dimconstraint$ eigenvectors associated with them forming the basis of the null space of $\samplecovariancematrix$. On the other hand, the remaining $(\dimambient - \dimconstraint)$ eigenvectors form the basis of the row space of $\samplecovariancematrix$. 
We follow the same technique as \citet{Deutsch2015_TensorVotingGraph} for automatically determining the number of constraints $\dimconstraint$ from data, which is also the number of outputs of {\ecmnn}\footnote{Here we assume that the intrinsic dimensionality of the manifold at each point remains constant.}. 
Suppose the eigenvalues of $\samplecovariancematrix$ are $\{\coveigval_1, \coveigval_2, \dots, \coveigval_\dimambient\}$ (in decreasing order w.r.t. magnitude). Then the number of constraints can be determined as $\dimconstraint = \argmax{\left(\left[\coveigval_1 - \coveigval_2, \coveigval_2 - \coveigval_3, \dots, \coveigval_{\dimambient-1} - \coveigval_{\dimambient}
\right]\right)}$.

We now present several methods to define and train \ecmnn, as follows:

\subsection{Alignment of local tangent and normal spaces}
{\ecmnn} aims to align the following:
\begin{enumerate}[(a), leftmargin=*, itemsep=0em, topsep=0em]
    \item the null space of $\constraintmanifoldjacobian$ and the row space of $\samplecovariancematrix$ (which must be equivalent to tangent space $\tangentspaceatjointposition$)
    \item the row space of $\constraintmanifoldjacobian$ and the null space of $\samplecovariancematrix$ (which must be equivalent to normal space $\normalspaceatjointposition$)
\end{enumerate}
for the local neighborhood of each point $\jointposition$ in the on-manifold dataset. 
Suppose the eigenvectors of $\samplecovariancematrix$ are $\{\coveigvec_1, \coveigvec_2, \dots, \coveigvec_\dimambient\}$ and the singular vectors of $\constraintmanifoldjacobian$ are $\{\constrjaceigvec_1, \constrjaceigvec_2, \dots, \constrjaceigvec_\dimambient\}$, where the indices indicate decreasing order w.r.t. the eigenvalue/singular value magnitude. The null spaces of $\samplecovariancematrix$ and $\constraintmanifoldjacobian$ are spanned by $\{\coveigvec_{\dimambient-\dimconstraint+1}, \dots, \coveigvec_\dimambient\}$ and $\{\constrjaceigvec_{\dimconstraint+1}, \dots, \constrjaceigvec_\dimambient\}$, respectively. The two conditions above imply that the projection of the null space eigenvectors of $\constraintmanifoldjacobian$ into the null space of $\samplecovariancematrix$ should be $\vct{0}$, and similarly for the row spaces. 
Hence, we achieve this by training {\ecmnn} to minimize projection errors $\norm{\covnullspaceeigmat \covnullspaceeigmat\T \constrjacnullspaceeigmat}_2^2$ and $\norm{ \constrjacnullspaceeigmat \constrjacnullspaceeigmat\T \covnullspaceeigmat}_2^2$ with 
$\covnullspaceeigmat = 
\begin{bmatrix}
    \coveigvec_{\dimambient-\dimconstraint+1} & \dots & \coveigvec_\dimambient
\end{bmatrix}$
and 
$\constrjacnullspaceeigmat =  
\begin{bmatrix}
    \constrjaceigvec_{\dimconstraint+1} & \dots & \constrjaceigvec_\dimambient
\end{bmatrix}$ \new{at all the points in the on-manifold dataset}.

\subsection{Data augmentation with off-manifold data}
\label{ssec:data_augmentation}
The training dataset is on-manifold, i.e., each point $\jointposition$ in the dataset satisfies $\constraintfunction(\jointposition) = \vct{0}$. 
Through Local PCA on each of these points, we know the data-driven approximation of the normal space of the manifold at $\jointposition$. 
Hence, we know the directions where the violation of the equality constraint increases, i.e., the same direction as any vector picked from the approximate normal space. 
Since our future use of the learned constraint manifold on motion planning does not require the acquisition of the near-ground-truth value of $\constraintfunction(\jointposition) \neq \vct{0}$, we can set this off-manifold valuation of $\constraintfunction$ arbitrarily, as long as it does not interfere with the utility for projecting an off-manifold point onto the manifold. 
Therefore, we can augment our dataset with off-manifold data to achieve a more robust learning of {\ecmnn}. 
For each point $\jointposition$ in the on-manifold dataset, and for each random unit vector $\randunitvec$ picked from the normal space at $\jointposition$, we can add an off-manifold point $\augjointposition = \jointposition + \augidx \augmagnitude \randunitvec$ with a positive integer $\augidx$ and 
a small positive scalar $\augmagnitude$ (see Figure \ref{fig:localpca} for a visualization). 
However, if the
closest on-manifold data point to an augmented point
$\augjointposition = \jointposition + \augidx \augmagnitude \randunitvec$ 
is not $\jointposition$, we reject it.
This prevents situations like augmenting a point on a sphere beyond the center of the sphere. 
\new{Data augmentation is a technique used in various fields, and our approach has similarities to others \citep{chin2008out, bellinger2018manifold, patel2019manifold}, 
though in this work we focus on using augmentation to aid learning an implicit constraint function for robotic motion planning.}
With this data augmentation, we now define several losses used to train {\ecmnn}.
\subsubsection{Training losses}
\label{sssec:losses}
\textbf{Loss based on the norm of the output vector of {\ecmnn}} 
For the augmented data point $\augjointposition = \jointposition + \augidx \augmagnitude \randunitvec$, we set the label satisfying $\norm{\constraintfunction(\augjointposition)}_2 = \augidx \augmagnitude$. During training, we minimize the norm prediction error $\normloss = \norm{(\norm{\constraintfunction(\augjointposition)}_2 - \augidx \augmagnitude)}_2^2$ for each augmented point $\augjointposition$.
\\
\new{Furthermore, we define the following three siamese losses. The main intuition behind these losses is that we expect the learned function $\constraintfunction$ to output similar values for similar points.\\}
\textbf{Siamese loss for reflection pairs}
For the augmented data point $\augjointposition = \jointposition + \augidx \augmagnitude \randunitvec$ and its reflection pair $\jointposition - \augidx \augmagnitude \randunitvec$, we can expect that $\constraintfunction(\jointposition + \augidx \augmagnitude \randunitvec) = -\constraintfunction(\jointposition - \augidx \augmagnitude \randunitvec)$\new{, 
or in other words, that an augmented point and its reflection pair should have the same $\constraintfunction$ but with opposite signs}.
Therefore, during training we minimize the siamese loss $\siamreflectionloss = \norm{\constraintfunction(\jointposition + \augidx \augmagnitude \randunitvec) + \constraintfunction(\jointposition - \augidx \augmagnitude \randunitvec)}_2^2$.
\\
\textbf{Siamese loss for augmentation fraction pairs}
Similarly, between the pair $\augjointposition = \jointposition + \augidx \augmagnitude \randunitvec$ and $\jointposition + \frac{a}{b} \augidx \augmagnitude \randunitvec$, where $a$ and $b$ are positive integers satisfying $0 < \frac{a}{b} < 1$, we can expect that $\frac{\constraintfunction(\jointposition + \augidx \augmagnitude \randunitvec)}{\norm{\constraintfunction(\jointposition + \augidx \augmagnitude \randunitvec)}_2} = \frac{\constraintfunction(\jointposition + \frac{a}{b} \augidx \augmagnitude \randunitvec)}{\norm{\constraintfunction(\jointposition + \frac{a}{b} \augidx \augmagnitude \randunitvec)}_2}$. 
\new{In other words, the \textit{normalized} $\constraintfunction$ values should be the same on an augmented point $\jointposition + \augidx \augmagnitude \randunitvec$ as on any point in between the on-manifold point $\jointposition$ and that augmented point $\jointposition + \augidx \augmagnitude \randunitvec$.}
Hence, during training we minimize the siamese loss \mbox{$\siamfracloss = \Big|\Big|{\frac{\constraintfunction(\jointposition + \augidx \augmagnitude \randunitvec)}{\norm{\constraintfunction(\jointposition + \augidx \augmagnitude \randunitvec)}_2} - \frac{\constraintfunction(\jointposition + \frac{a}{b} \augidx \augmagnitude \randunitvec)}{\norm{\constraintfunction(\jointposition + \frac{a}{b} \augidx \augmagnitude \randunitvec)}_2}}\Big|\Big|_2^2$.}
\\
\textbf{Siamese loss for similar augmentation pairs}
Suppose for nearby on-manifold data points $\jointposition_a$ and $\jointposition_c$, their approximate normal spaces $\normalspaceid_{\jointposition_a}\constraintmanifold$ and $\normalspaceid_{\jointposition_c}\constraintmanifold$ are spanned by eigenvector bases $\lnormalcoordframe^a = \{\coveigvec^a_{\dimambient-\dimconstraint+1}, \dots, \coveigvec^a_\dimambient\}$ and $\lnormalcoordframe^c = \{\coveigvec^c_{\dimambient-\dimconstraint+1}, \dots, \coveigvec^c_\dimambient\}$, respectively.
If $\lnormalcoordframe^a$ and $\lnormalcoordframe^c$ are closely aligned, the random unit vectors
$\randunitvec_a$ from $\lnormalcoordframe^a$ and $\randunitvec_c$ from $\lnormalcoordframe^c$ can be obtained by 
$\randunitvec_a = \frac{\sum_{j=\dimambient-\dimconstraint+1}^\dimambient \randscalarweight_j \coveigvec^a_j}{\norm{\sum_{j=\dimambient-\dimconstraint+1}^\dimambient \randscalarweight_j \coveigvec^a_j}_2}$ 
and $\randunitvec_c = \frac{\sum_{j=\dimambient-\dimconstraint+1}^\dimambient \randscalarweight_j \coveigvec^c_j}{\norm{\sum_{j=\dimambient-\dimconstraint+1}^\dimambient \randscalarweight_j \coveigvec^c_j}_2}$, 
where $\{\randscalarweight_{\dimambient-\dimconstraint+1}, \dots, \randscalarweight_\dimambient\}$ 
are random scalar weights from a standard normal distribution common to both the bases of $\lnormalcoordframe^a$ and $\lnormalcoordframe^c$. This will ensure that $\randunitvec_a$ and $\randunitvec_c$ are aligned as well, and we can expect that  $\constraintfunction(\jointposition_a + \augidx \augmagnitude \randunitvec_a) = \constraintfunction(\jointposition_c + \augidx \augmagnitude \randunitvec_c)$. 
\new{In other words, two aligned augmented points in the same level set should have the same $\constraintfunction$ value.} 
Therefore, during training we minimize the siamese loss $\siamsimilarloss = \norm{\constraintfunction(\jointposition_a + \augidx \augmagnitude \randunitvec_a) - \constraintfunction(\jointposition_c + \augidx \augmagnitude \randunitvec_c)}_2^2$. 
In general, the alignment of $\lnormalcoordframe^a$ and $\lnormalcoordframe^c$ is not guaranteed, 
for example due to the numerical sensitivity of singular value/eigen decomposition. 
Therefore, we introduce an algorithm for Orthogonal Subspace Alignment (OSA) in the Supplementary Material to ensure that this assumption is satisfied.

While $\normloss$ governs only the norm of {\ecmnn}'s output, the other three losses $\siamreflectionloss$, $\siamfracloss$, and $\siamsimilarloss$ constrain the (vector) outputs of {\ecmnn} based on pairwise input data points without explicitly hand-coding the desired output itself. We avoid the hand-coding of the desired output because this is difficult for high-dimensional manifolds, except when there is prior knowledge about the manifold available, such as in the case of Signed Distance Fields (SDF) manifolds.

\section{Experiments}
\label{sec:experiments}
\begin{figure}[t]
\begin{subfigure}{0.31\textwidth}
\includegraphics[trim={1cm 1cm 1cm 1cm}, clip, height=0.8\textwidth]
{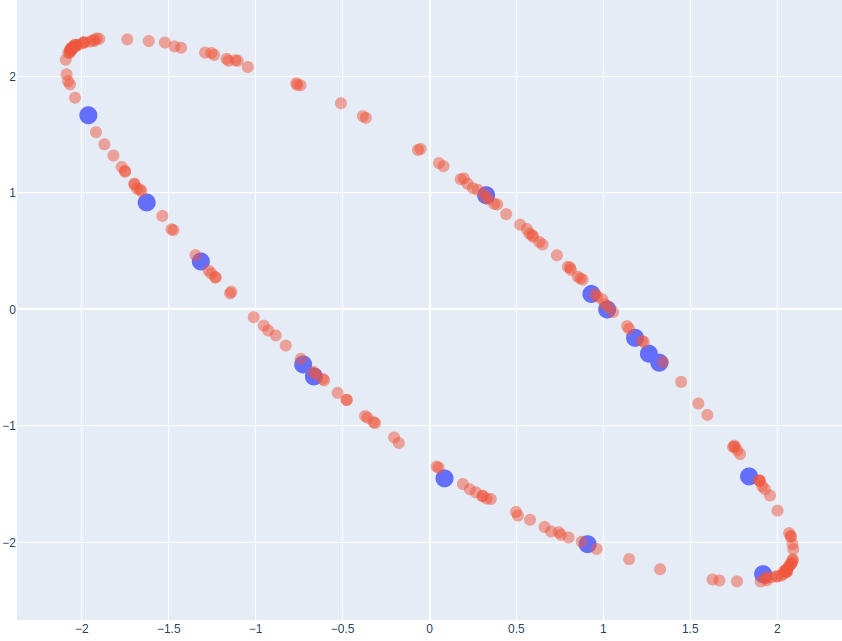}
\caption{Samples from {\ecmnn}}
\label{fig:ecomann_3dof_sample_a}
\end{subfigure}
\hspace{.2em}
\begin{subfigure}{0.31\textwidth}
\includegraphics[trim={1cm 1cm 1cm 1cm}, clip, height=0.8\textwidth]
{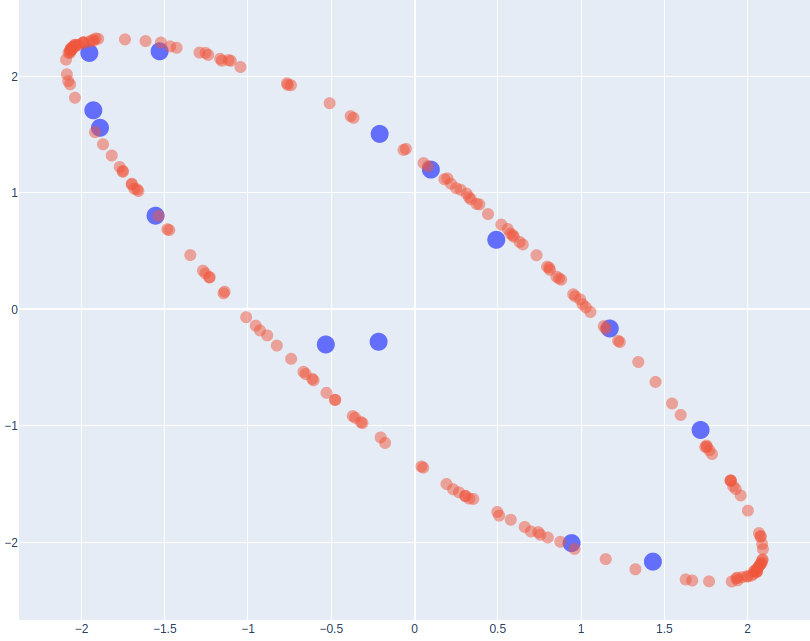}
\caption{Samples from VAE}
\label{fig:vae_3dof_sample_a}
\end{subfigure}
\hspace{.2em}
\begin{subfigure}{0.31\textwidth}
    \centering
    \includegraphics[height=0.8\textwidth]{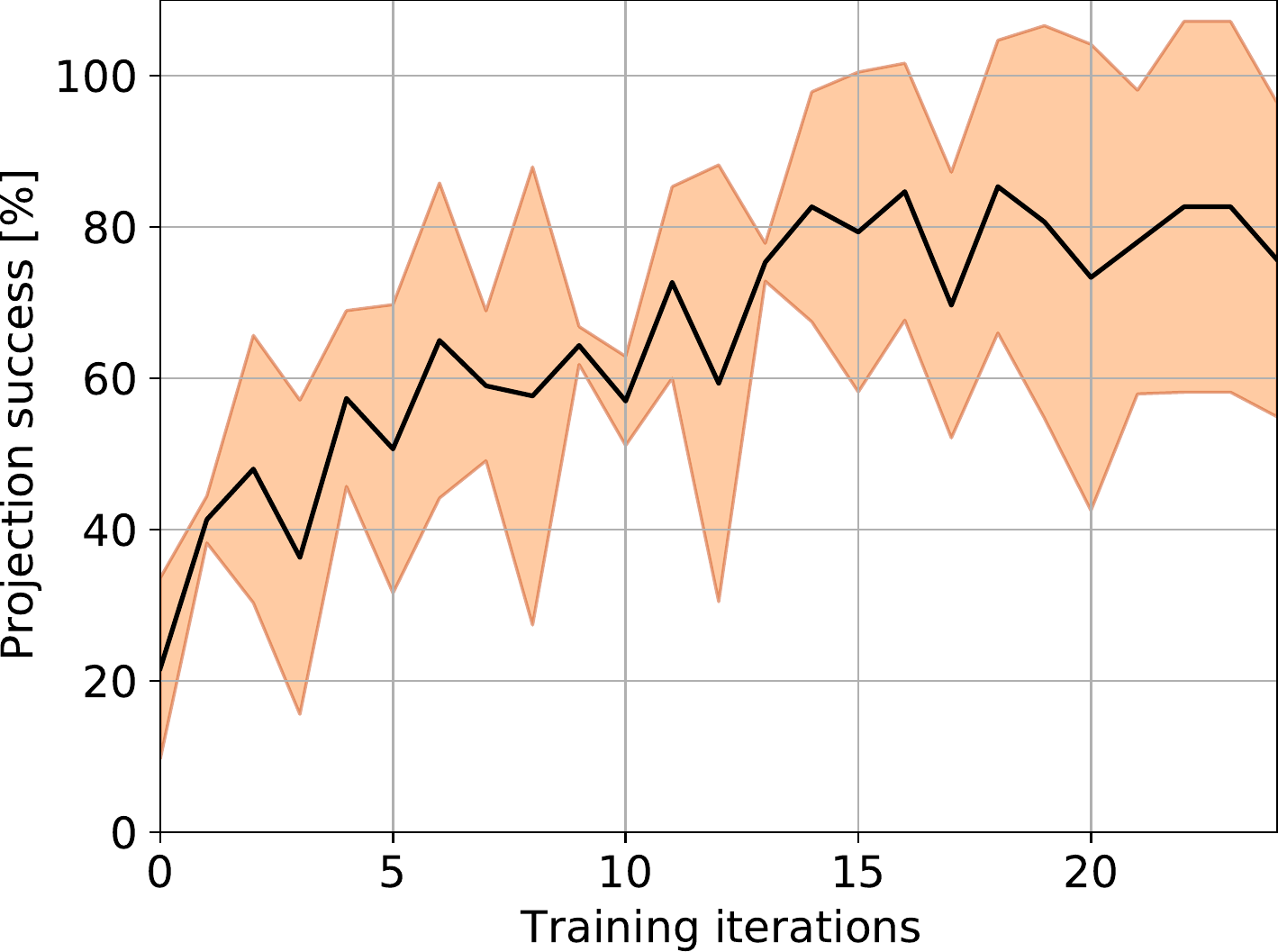}
    \caption{{\ecmnn} projection success}
    \label{fig:3dof_success_rate}
\end{subfigure}
\caption{Images \subref{fig:ecomann_3dof_sample_a} and \subref{fig:vae_3dof_sample_a} 
visualize a slice near $z=0$ of the Plane dataset for experiment \ref{sec:experiment1}. Red points are the training dataset and blue points are samples generated from the learned manifolds.
The points projected onto the manifold using {\ecmnn} are closer to the manifold, with an 85\% projection success rate. 
A significant portion of the points generated using the VAE lie inside the surface, 
which leads to a lower success rate of 77\%. Figure \subref{fig:3dof_success_rate} shows the projection success of {\ecmnn} over the number of training iterations.
The quantitative results are found in Table \ref{table:projection}.}
\label{fig:samples}
\end{figure}

We use the robot simulator MuJoCo \citep{todorov2012mujoco} to generate four datasets. The size of each dataset is denoted as $\numdatapoints$. We define a ground truth constraint $\bar{h}_M$, randomly sample points in the configuration (joint) space, and use a constrained motion planner to find robot configurations in $\onconstraintconfigspace$ that produce the on-manifold datasets:
    \textbf{Sphere}: 3D point that has to stay on the surface of a sphere. $\numdatapoints=5000, \dimambient=3, \dimconstraint=1$.
    \textbf{3D Circle}: A point that has to stay on a circle in 3D space. $\numdatapoints=1000, \dimambient=3, \dimconstraint=2$.
    \textbf{Plane}: Robot arm with 3 rotational DoFs where the end effector has to be on a plane. $\numdatapoints=20000, \dimambient=3, \dimconstraint=1$.
    \textbf{Orient}: Robot arm with 6 rotational DoFs that has to keep its orientation upright (e.g., transporting a cup). $\numdatapoints=21153, \dimambient=6, \dimconstraint=2$. \new{In the following experiments, we parametrize {\ecmnn} with $4$ hidden layers of size $36$, $24$, $18$, and $10$. The hidden layers use a $\tanh$ activation function and the output layer is linear.}

\subsection{Accuracy and precision of learned manifolds} 
\label{sec:experiment1}
\begin{table}[t]
  \caption{Accuracy and precision of learned manifolds averaged across 3 instances. 
 ``Train" indicates results on the on-manifold training set;  ``test" indicates $\numdatapoints=1000$ projected (\ecmnn) or sampled (VAE) points.}
  \label{table:projection}
  \centering
  \resizebox{\columnwidth}{!}{
\begin{tabular}{c||ccc|ccc}
    \toprule
      & \multicolumn{3}{c}{\ecmnn} & \multicolumn{3}{c}{VAE} \\
     \cmidrule{2-7}
              Dataset & $\mu_{\bar{h}_M}$ (train) & $\mu_{\bar{h}_M}$ (test) & $P_{\bar{h}_M}$
              & $\mu_{\bar{h}_M}$ (train) & $\mu_{\bar{h}_M}$ (test) & $P_{\bar{h}_M}$
                \\
\midrule
      Sphere 
      & $0.024\pm 0.009$ & $0.023 \pm 0.009$ & $100.0 \pm 0.0$ 
      & $0.105\pm 0.088$ & $0.161 \pm 0.165$ & $46.867 \pm 18.008$
      \\
      3D Circle     
      & $0.029 \pm 0.011$ & $0.030 \pm 0.011$ & $78.0 \pm 22.0$ 
        & $0.894 \pm 0.074$ & $0.902 \pm 0.069$ & $0.0 \pm 0.0$   \\
      Plane  
      & $0.020 \pm 0.005$ & $0.020 \pm 0.005$ & $88.5 \pm 10.5$ 
        & $0.053 \pm 0.075$ & $0.112 \pm 0.216$  & $77.733 \pm 7.721$ \\
      Orient     & $0.090\pm 0.009$ & $0.090 \pm 0.009$ & $73.5 \pm 6.5$
        & $0.010 \pm 0.037$ & $0.085 \pm 0.237$ & $85.9 \pm 1.068$   \\
      \bottomrule
  \end{tabular}
  }
\end{table}
We compare the accuracy and precision of the manifolds learned by {\ecmnn}
with those learned by a variational autoencoder (VAE) \citep{kingma2013AutoEncodingVB}. 
VAEs are a popular generative model that embeds data points as a distribution in a learned latent space, and as such new latent vectors
can be sampled and decoded into new examples which fit the distribution of
the training data.\footnote{We also tested classical manifold learning techniques (Isomap, LTSA, PCA, MDS, and LLE). We found them empirically not expressive enough and/or unable to support projection or sampling operations, necessary capabilities for this work.}
We use two metrics: First, the distance $\mu_{\bar{h}_M}$ which measures how far a point is away from the ground-truth manifold $\bar{h}_M$ and which we evaluate for both the training data points and randomly sampled points, and second,
the percent $P_{\bar{h}_M}$ of randomly sampled points that are on the manifold $\bar{h}_M$.
We use a distance threshold of $0.1$ to determine success when calculating $P_{\bar{h}_M}$.
For {\ecmnn}, randomly sampled points are projected using gradient descent with the learned implicit function until convergence. 
For the VAE, latent points are sampled from $\mathcal{N}(0,1)$ and decoded into new configurations.

\begin{figure}[t]
    \centering
    \begin{subfigure}{0.45\textwidth}
        \centering
        \includegraphics[trim={0cm 0cm 0cm 0cm}, clip, height=0.75\textwidth]{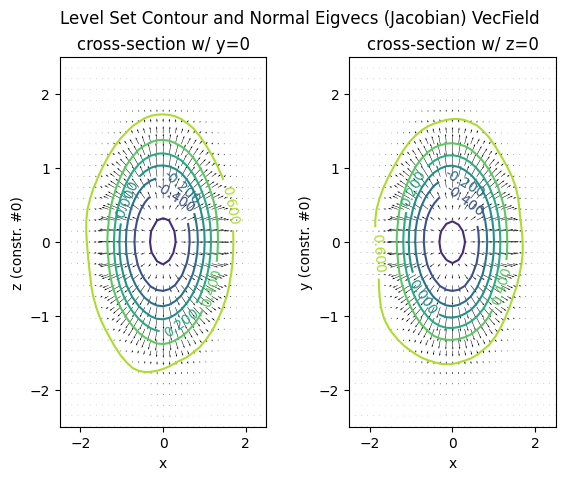}
        \label{sfig:contourplot_vecfield_sphere}
    \end{subfigure}
    \vspace{1em}
    \begin{subfigure}{0.45\textwidth}
        \centering
        \includegraphics[trim={0cm 0cm 0cm 0cm}, clip, height=0.75\textwidth]{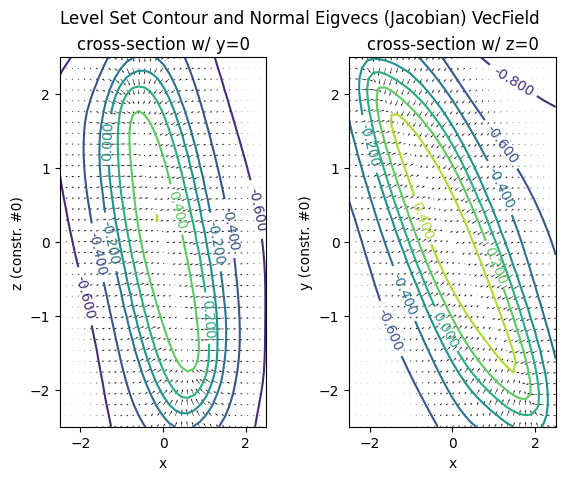}
        \label{sfig:contourplot_vecfield_3dof_traj}
    \end{subfigure}
    \caption{Trained {\ecmnn}’s level set contour plot and the normal space eigenvector field, after training on the sphere constraint dataset (left) and plane constraint dataset (right).}
    \label{fig:contourplot_vecfield}
\end{figure}

We sample 1000 points for each of these comparisons. 
We report results in Table \ref{table:projection} and a visualization of projected samples in Fig. \ref{fig:samples}. 
We also plot the level set and the normal space eigenvector field of the {\ecmnn} trained on the sphere and plane constraint dataset in Fig. \ref{fig:contourplot_vecfield}.
In all experiments, we set the value of the augmentation magnitude $\augmagnitude$ to the square root of the mean eigenvalues of the approximate tangent space, which we found to work well experimentally.
With the exceptions of the embedding size and the input size, which are set to the dimensionality $\dimambient-\dimconstraint$ as the tangent space of the constraint learned by {\ecmnn} 
and the ambient space dimensionality $\dimambient$ of the dataset, respectively,
the VAE has the same parameters for each dataset: 
4 hidden layers with 128, 64, 32, and 16 units in the encoder and the same but reversed in the decoder; the weight of the KL divergence loss $\beta = 0.01$; using batch normalization; and trained for 100 epochs.

Our results show that for every dataset except Orient, {\ecmnn} out performs the VAE in both metrics. 
{\ecmnn} additionally outperforms the VAE with the Orient dataset in the testing phase, which suggests more robustness of the learned model. 
We find that though the VAE also performs relatively well in most cases, it cannot learn a good representation of the 3D Circle constraint and fails to produce any valid sampled points. 
{\ecmnn}, on the other hand, can learn to represent all four constraints well.

\subsection{Ablation study of \ecmnn}
\begin{table}[t]
    \centering
    \caption{Percentage of projection success rate for a variety of ablations of {\ecmnn} components.}
    \begin{tabular}{ |c||c|c|c| }
        \hline
        Ablation Type         & Sphere                  & 3D Circle               & Plane \\
        \hline
        \hline
        No Ablation           & (98.67 $\pm$ 1.89) \% & (100.00 $\pm$  0.00) \% & (92.33 $\pm$ 10.84) \% \\
        w/o Data Augmentation & ( 9.00 $\pm$ 2.45) \% & ( 16.67 $\pm$  4.64) \% & ( 3.67 $\pm$  3.30) \% \\
        w/o OSA               & (64.33 $\pm$ 9.03) \% & ( 33.33 $\pm$ 17.13) \% & (61.00 $\pm$  6.16) \% \\
        w/o Siamese Losses    & (17.33 $\pm$ 3.40) \% & ( 65.67 $\pm$ 26.03) \% & (35.67 $\pm$  5.56) \% \\
        w/o Siamese Loss $\siamreflectionloss$ & (92.67 $\pm$ 4.03) \% & (  9.67 $\pm$  2.05) \% & (38.00 $\pm$ 16.87) \% \\
        w/o Siamese Loss $\siamfracloss$       & (88.33 $\pm$ 8.34) \% & ( 99.67 $\pm$  0.47) \% & (85.33 $\pm$ 16.68) \% \\
        w/o Siamese Loss $\siamsimilarloss$    & (83.00 $\pm$ 5.10) \% & ( 70.67 $\pm$ 21.00) \% & (64.33 $\pm$  2.62) \% \\
        \hline
    \end{tabular}
    \label{tab:ablation_table}
\end{table}
In the ablation study, we compare $P_{\bar{h}_M}$ across 7 different {\ecmnn} setups: 1) no ablation;
2) without data augmentation; 3) without orthogonal subspace alignment (OSA) during data augmentation; 4) without siamese losses during training\new{; 5) without $\siamreflectionloss$; 6) without $\siamfracloss$; and 7) without $\siamsimilarloss$}. Results are reported in Table \ref{tab:ablation_table}. 
The data suggest that all parts of the training process are essential for a high success rate during projection. 
Of the features tested, data augmentation appears to have the most impact.
This makes sense because without augmented data to train on, any configuration that does not already lie on the manifold will have undefined output when evaluated with {\ecmnn}. 
\new{Additionally, results from ablating the individual siamese losses suggest that the contribution of each is dependent on the context and structure of the constraint.
Complementary to this ablation study, we present some additional experimental results in the Supplementary Material.}

\subsection{Motion planning on learned manifolds}
\begin{wrapfigure}{r}{0.26\textwidth}
    \centering
     \vspace{-10pt}
    \includegraphics[width=0.26\textwidth]{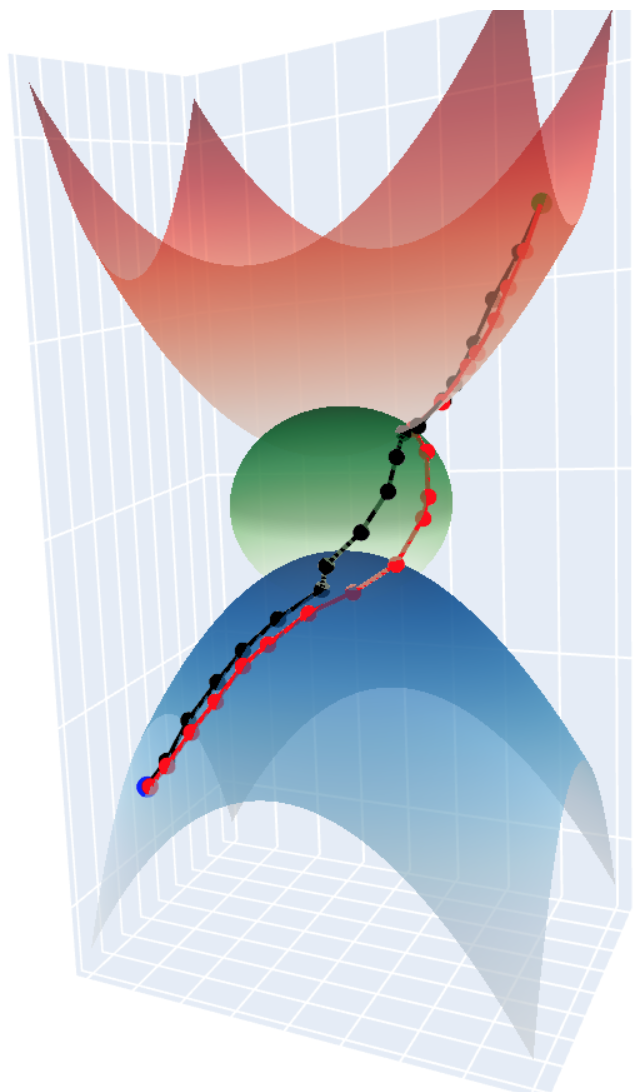}
    \caption{Planned path on the learned manifold (red) and on the ground truth manifold (black).}
    \vspace{-20pt}
    \label{fig:hourglass}
\end{wrapfigure}
In the final experiment, we integrate {\ecmnn} into the sequential motion planning framework described in Section \ref{subsec:smp}. 
We mix the learned constraints with analytically defined constraints and evaluate it for two tasks. 
The first one is a geometric task, visualized in Figure \ref{fig:hourglass}, where a point starting on a paraboloid in 3D space must find a path to a goal state on another paraboloid. 
The paraboloids are connected by a sphere, and the point is constrained to stay on the surfaces at all times.
In this case, we give the paraboloids analytically to the planner, and use {\ecmnn} to learn the connecting constraint using the Sphere dataset.
Figure \ref{fig:hourglass} shows the resulting path where the sphere is represented by a learned manifold (red line) and where it is represented by the ground-truth manifold (black line). While the paths do not match exactly, both paths are on the manifold and lead to similar solutions in terms of path lengths. 
\new{The task was solved in \SI{27.09}{\s} on a 2.2 GHz Intel Core i7 processor. The tree explored $1117$ nodes and the found path consists of $24$ nodes.}

\begin{figure}[t]
    \centering
    \includegraphics[width=0.18\textwidth]{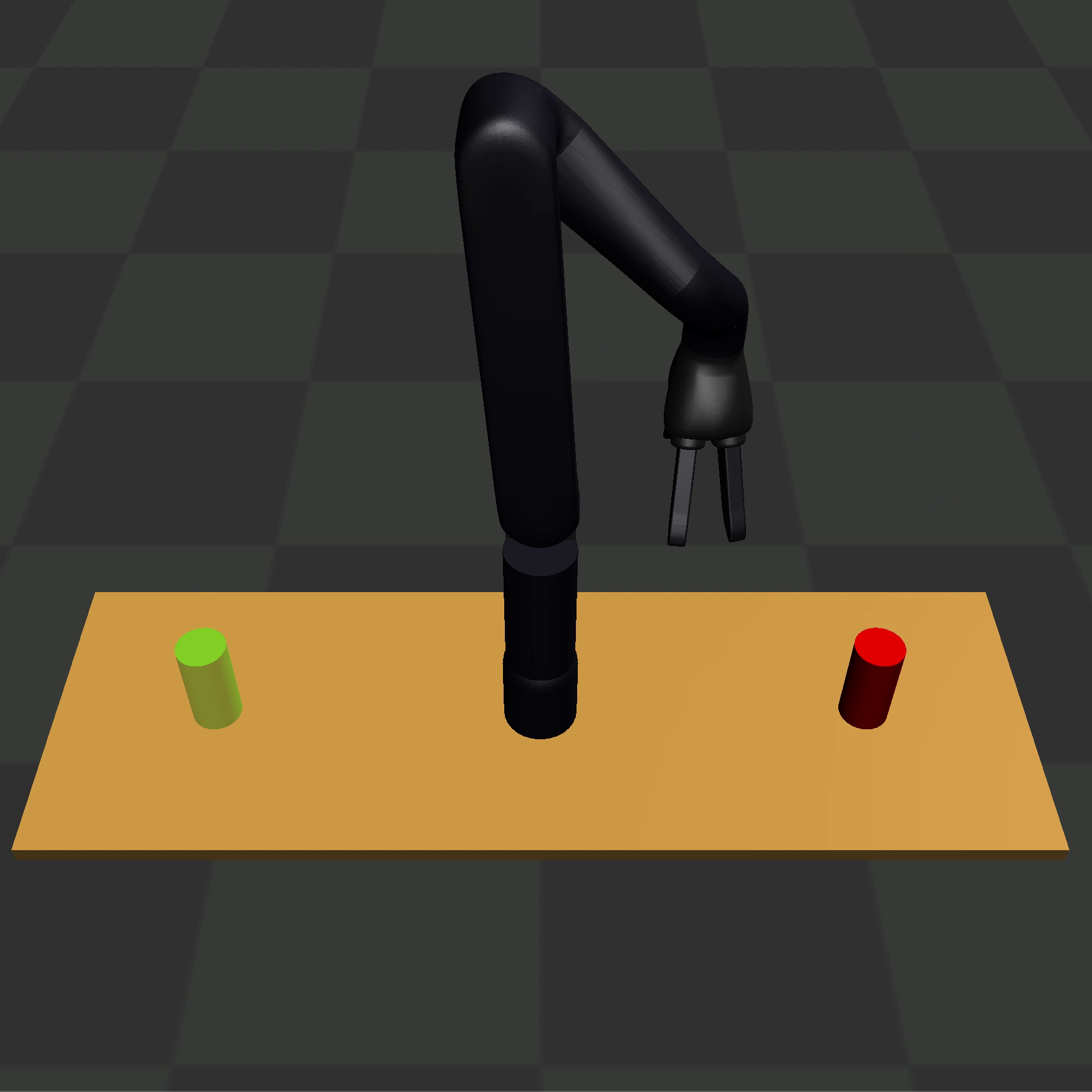}
    \includegraphics[width=0.18\textwidth]{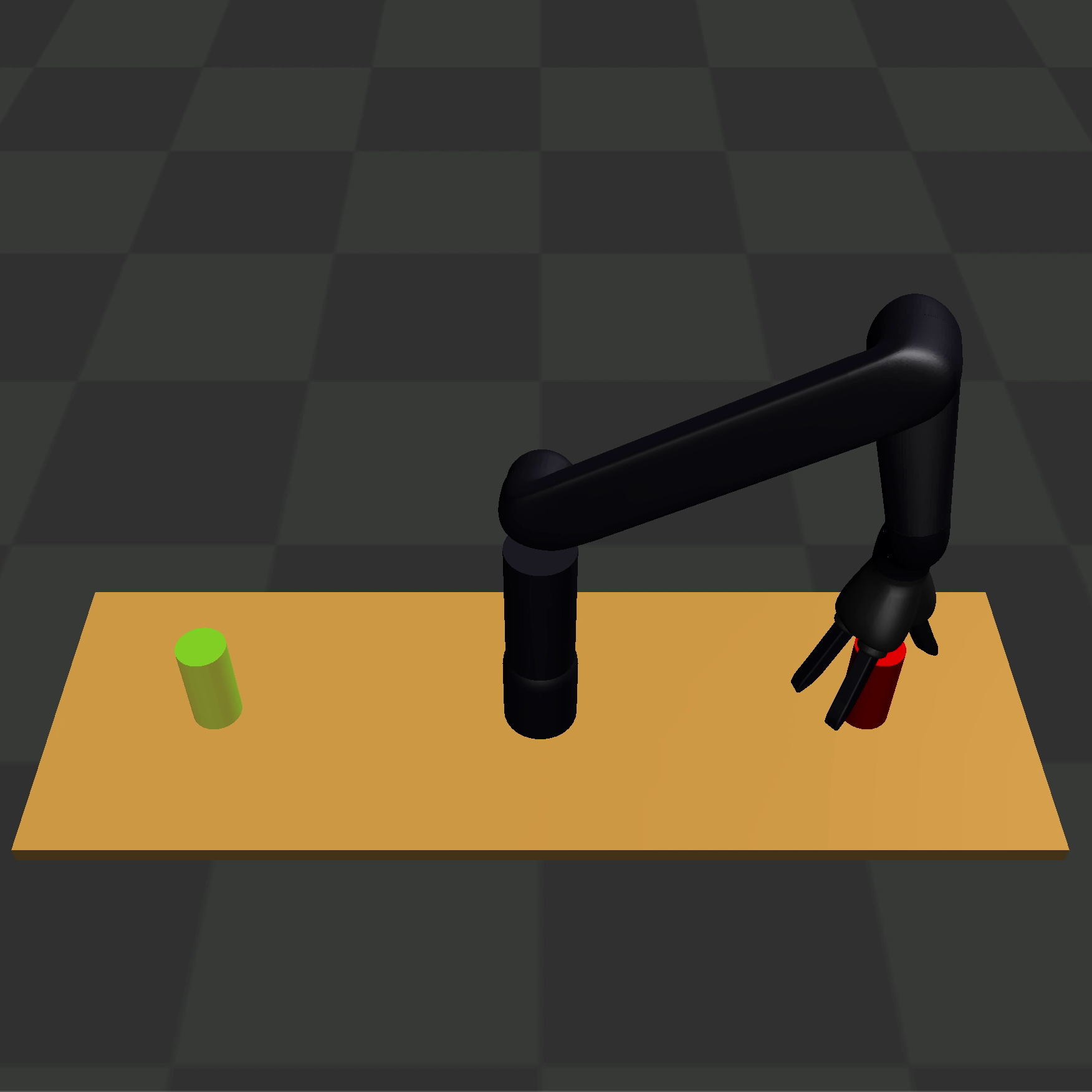}
    \includegraphics[width=0.18\textwidth]{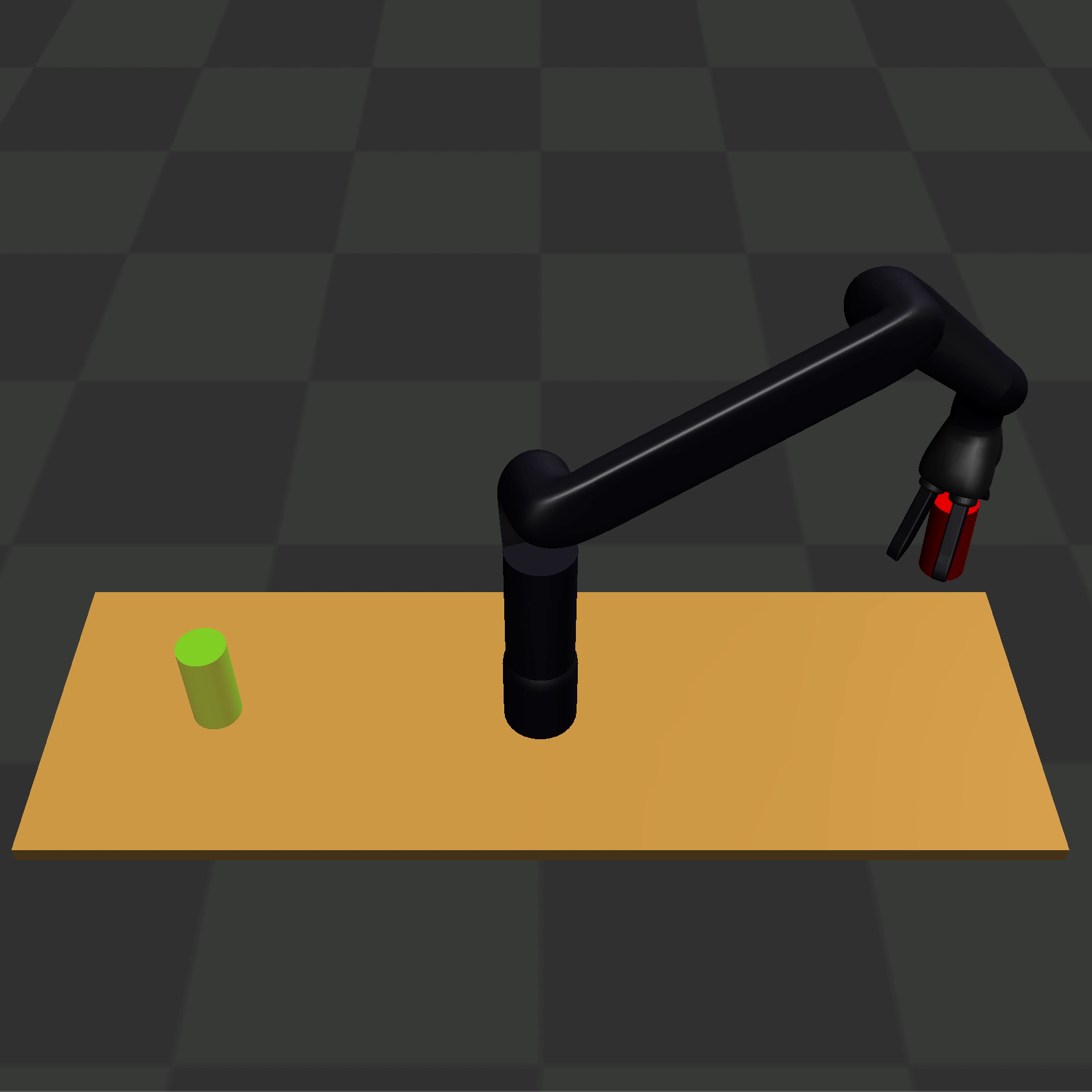}
    \includegraphics[width=0.18\textwidth]{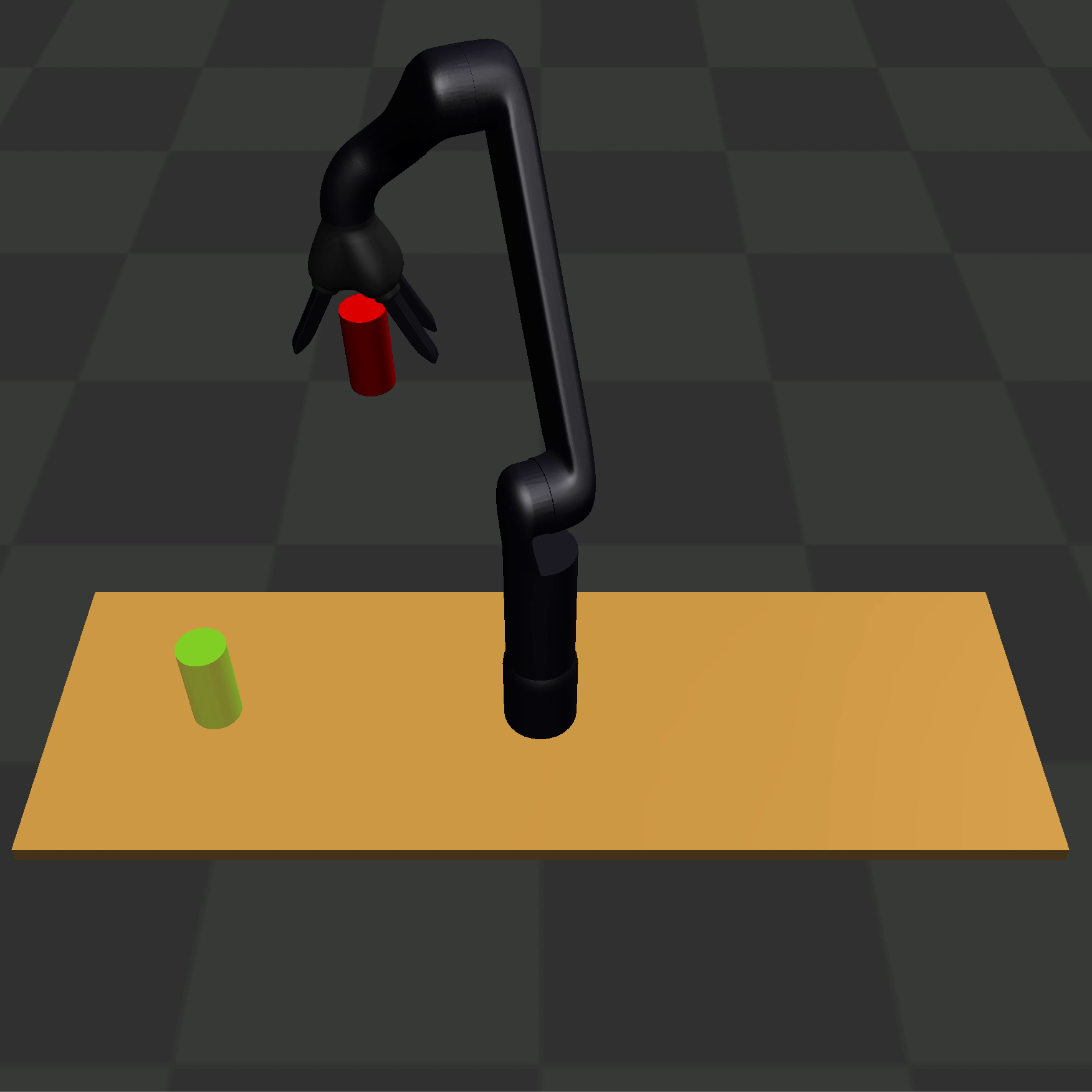}
    \includegraphics[width=0.18\textwidth]{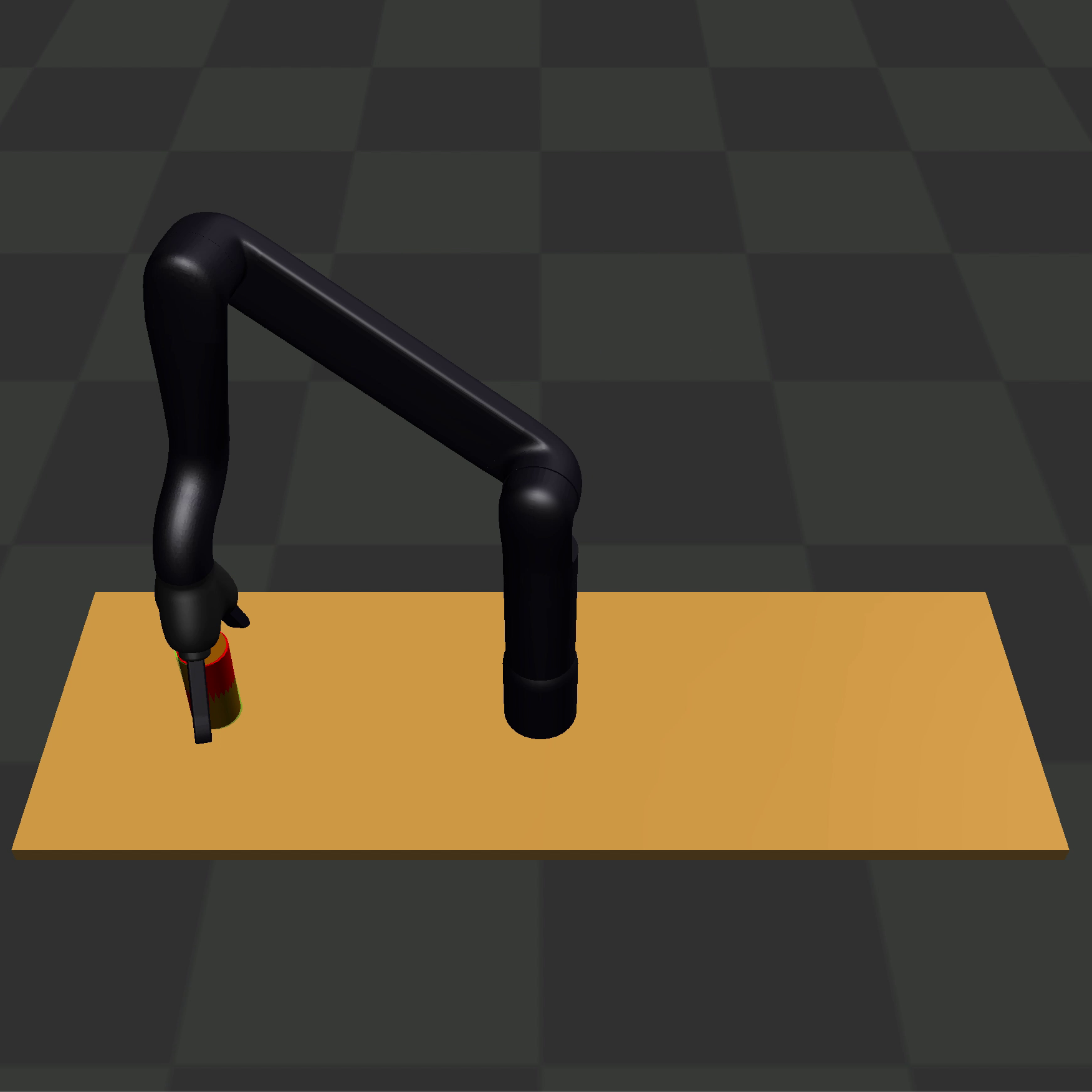}
    \caption{The images visualize a path that was planned on a learned orientation manifold.}
    \label{fig:pickandplace}
\end{figure}

The second task is a robot pick-and-place task with the additional constraint that the transported object needs to be oriented upwards throughout the whole motion. 
For this, we use the Orient dataset to learn the manifold for the transport phase and combine it with other manifolds that describe the pick and place operation. 
\new{The planning time was \SI{42.97}{\s}, the tree contained $1421$ nodes and the optimal path had $22$ nodes.} 
Images of the resulting path are shown in Figure \ref{fig:pickandplace}. 

\section{Discussion and conclusion}
\label{sec:conclusion}
In this paper, we presented a novel method called {\ecmnnlong} for learning equality constraint manifolds from data.
{\ecmnn} works by aligning the row and null spaces of the local PCA and network Jacobian, which results in approximate learned normal and tangent spaces of the underlying manifold, suitable for use within a constrained sampling-based motion planner.
In addition, we introduced a method for augmenting a purely on-manifold dataset to include off-manifold points and several loss functions for training.
This improves the robustness of the learned method while avoiding hand-coding the labels for the augmented points. We also showed that the learned manifolds can be used in a sequential motion planning framework for constrained robot tasks.

While our experiments show success in learning a variety of manifolds, there are some limitations to our method.
First, {\ecmnn} by design can only learn equality constraints. Although many tasks can be specified with such constraints, inequality constraints are also an important part of many robot planning problems. 
Additionally, because of inherent limitations in learning from data, {\ecmnn} does not guarantee that a randomly sampled point in configuration space will be projected successfully onto the learned manifold.
This presents challenges in designing asymptotically complete motion planning algorithms, and is an important area of research.
In the future, we plan on further testing {\ecmnn} on more complex tasks, and in particular on tasks which are demonstrated by a human rather than from simulation.

\clearpage

\acknowledgments{This material is based upon work supported by the National Science Foundation Graduate Research Fellowship Program under Grant No. DGE-1842487. Any opinions, findings, and conclusions or recommendations expressed in this material are those of the author(s) and do not necessarily reflect the views of the National Science Foundation.
This work was supported in part by the Office of Naval Research (ONR) under grant N000141512550.}

\bibliography{bibliography}

\newpage

\appendix

\section{Supplementary Materials}
We provide the following supplementary material to enhance the main paper:
\begin{itemize}
    \item \textbf{Orthogonal Subspace Alignment} -- In Section \ref{sec:osa}, we describe the details of the orthogonal subspace alignment and provide an algorithm that shows its step-by-step computations.
    \new{\item \textbf{Additional Experiments} -- In order to thoroughly evaluate the components of our approach, we present some additional experimental results in Section \ref{sec:addtnl_exp}.}
\end{itemize}
\label{sec:osa}
\subsection{Orthogonal Subspace Alignment (OSA)}
\label{sec:osa}
In previous work \cite{thopalli2019subspacealignment, he2020quantum}, subspace alignment techniques -- without orthogonality constraints -- have been introduced to improve domain adaptation. For the purposes of this paper, we require a subspace alignment algorithm that preserves orthogonality of the subspaces being aligned, which we present in this section.

Given a set of orthonormal vectors $\coordframe = \{\orthobasisvec_1, \orthobasisvec_2, \dots, \orthobasisvec_\dimambient\}$ which spans a \emph{space}, the matrix  $\orthobasismat = \begin{bmatrix}
    \orthobasisvec_1 & \orthobasisvec_2 & \dots & \orthobasisvec_\dimambient
\end{bmatrix} \in \Re^{\dimambient \times \dimambient}$ belongs to the Orthogonal Group $O(\dimambient)$.
The Orthogonal Group has two connected components, where one connected component called the Special Orthogonal Group $SO(\dimambient)$ is characterized by determinant $1$, and the other is characterized by determinant $-1$.  
However, if $\orthobasismat = 
\begin{bmatrix}
    \orthobasisvec_1 & \orthobasisvec_2 & \dots & \orthobasisvec_\dimambient
\end{bmatrix}$ has determinant 1 (i.e. if $\orthobasismat \in SO(\dimambient)$), 
then substituting $\orthobasisvec_1$ with its additive inverse ($-\orthobasisvec_1$)  will result in $\flippedorthobasismat = 
\begin{bmatrix}
    -\orthobasisvec_1 & \orthobasisvec_2 & \dots & \orthobasisvec_\dimambient
\end{bmatrix}$ with determinant $-1$. Aligning two coordinate frames  $\coordframe^a$ and $\coordframe^c$ to have a common origin and associated basis matrices $\orthobasismat_a$ and $\orthobasismat_c$, respectively, is equivalent to finding an $\rotmat \in SO(\dimambient)$ such that $\orthobasismat_a \rotmat = \orthobasismat_c$. The solution to this problem exists if and only if $\orthobasismat_a$ and $\orthobasismat_c$ come from the same connected component of $O(\dimambient)$, i.e. if either both $\orthobasismat_a, \orthobasismat_c \in SO(\dimambient)$ or both determinants of $\orthobasismat_a$ and $\orthobasismat_c$ are $-1$.

For a \emph{subspace} such as the normal space $\normalspaceatjointposition$ associated with an on-manifold data point $\jointposition$ on $M$ spanned by the eigenvectors $\lnormalcoordframe = \{\coveigvec_{\dimambient-\dimconstraint+1}, \dots, \coveigvec_\dimambient\}$, the concept of a determinant does not apply to $\covnullspaceeigmat = 
\begin{bmatrix}
    \coveigvec_{\dimambient-\dimconstraint+1} & \dots & \coveigvec_\dimambient
\end{bmatrix} \in \Re^{\dimambient \times \dimconstraint}$, as it is not a square matrix.
However, the normal space $\normalspaceatjointposition$ can be described with infinitely-many orthonormal bases ${\covnullspaceeigmat}_0$, ${\covnullspaceeigmat}_1$, ${\covnullspaceeigmat}_2$, ... ${\covnullspaceeigmat}_\infty$ where the set of column vectors of each is an orthonormal basis of $\normalspaceatjointposition$. Each of these is a member of $\Re^{\dimambient \times \dimconstraint}$. 
Moreover, we can pick the transpose of one of them, for example ${\covnullspaceeigmat}_0\T$, as a projection matrix, and ${\covnullspaceeigmat}_0$ as the inverse projection matrix. 
Applying the projection operation to each of the orthonormal bases, we get
${\covnullspaceolmat}_0 = {\covnullspaceeigmat}_0\T{\covnullspaceeigmat}_0 = \eye_{\dimconstraint \times \dimconstraint}$, ${\covnullspaceolmat}_1 = {\covnullspaceeigmat}_0\T{\covnullspaceeigmat}_1$, ${\covnullspaceolmat}_2 = {\covnullspaceeigmat}_0\T{\covnullspaceeigmat}_2$, ... ${\covnullspaceolmat}_\infty = {\covnullspaceeigmat}_0\T{\covnullspaceeigmat}_\infty$, and we will show that ${\covnullspaceolmat}_0$, ${\covnullspaceolmat}_1$, ${\covnullspaceolmat}_2$, ..., ${\covnullspaceolmat}_\infty$ are members of $O(\dimconstraint)$, which also has two connected components like $O(\dimambient)$. 
To show this, first note that although ${\covnullspaceeigmat}_0\T {\covnullspaceeigmat}_0 = \eye_{\dimconstraint \times \dimconstraint}$, the matrix ${\covnullspaceeigmat}_0 {\covnullspaceeigmat}_0\T \neq \eye_{\dimambient \times \dimambient}$. Hence, for any matrix $\mathbf{A} \in \Re^{\dimambient \times \dimambient}$ in general, ${\covnullspaceeigmat}_0 {\covnullspaceeigmat}_0\T \mathbf{A} \neq \mathbf{A}$.
However, we will show that ${\covnullspaceeigmat}_0 {\covnullspaceeigmat}_0\T \coveigvec = \coveigvec$ for any vector $\coveigvec$ in the vector space $\normalspaceatjointposition$. 
Suppose ${\covnullspaceeigmat}_0 = \begin{bmatrix}
    \orthobasisvec_1 & \orthobasisvec_2 & \dots & \orthobasisvec_\dimconstraint
\end{bmatrix} \in \Re^{\dimambient \times \dimconstraint}$, then we can write ${\covnullspaceeigmat}_0 {\covnullspaceeigmat}_0\T = \sum_{i=1}^{\dimconstraint} \orthobasisvec_i \orthobasisvec_i\T$.
Since the collection $\{\orthobasisvec_1, \orthobasisvec_2, \dots, \orthobasisvec_\dimconstraint\}$ spans the vector space $\normalspaceatjointposition$, any vector $\coveigvec$ in this vector space can be expressed as $\coveigvec = \sum_{i=1}^{\dimconstraint} \alpha_i \orthobasisvec_i$.
Moreover, $\orthobasisvec_i\T \coveigvec = \orthobasisvec_i\T \sum_{j=1}^{\dimconstraint} \alpha_j \orthobasisvec_j = \alpha_i$ for any $i = 1, 2, ..., \dimconstraint$, because by definition of orthonormality $\orthobasisvec_i\T \orthobasisvec_j = 1$ for $i = j$ and $\orthobasisvec_i\T \orthobasisvec_j = 0$ for $i \neq j$.
Hence, ${\covnullspaceeigmat}_0 {\covnullspaceeigmat}_0\T \coveigvec = (\sum_{i=1}^{\dimconstraint} \orthobasisvec_i \orthobasisvec_i\T) \coveigvec = \sum_{i=1}^{\dimconstraint} (\orthobasisvec_i\T \coveigvec) \orthobasisvec_i = \sum_{i=1}^{\dimconstraint} \alpha_i \orthobasisvec_i = \coveigvec$.
Similarly, because the column vectors of ${\covnullspaceeigmat}_0$, ${\covnullspaceeigmat}_1$, ${\covnullspaceeigmat}_2$, ..., ${\covnullspaceeigmat}_\infty$ are all inside the vector space $\normalspaceatjointposition$, it follows that ${\covnullspaceeigmat}_0 {\covnullspaceeigmat}_0\T {\covnullspaceeigmat}_0 = {\covnullspaceeigmat}_0$, ${\covnullspaceeigmat}_0 {\covnullspaceeigmat}_0\T {\covnullspaceeigmat}_1 = {\covnullspaceeigmat}_1$, ${\covnullspaceeigmat}_0 {\covnullspaceeigmat}_0\T {\covnullspaceeigmat}_2 = {\covnullspaceeigmat}_2$, ..., ${\covnullspaceeigmat}_0 {\covnullspaceeigmat}_0\T {\covnullspaceeigmat}_\infty = {\covnullspaceeigmat}_\infty$. 
Similarly, it can be shown that ${\covnullspaceeigmat}_i {\covnullspaceeigmat}_i\T \coveigvec = \coveigvec$ for any vector $\coveigvec$ in the vector space $\normalspaceatjointposition$ for any $i = 0, 1, 2, ..., \infty$. 
Furthermore, ${\covnullspaceolmat}_0\T {\covnullspaceolmat}_0 = {\covnullspaceeigmat}_0\T({\covnullspaceeigmat}_0{\covnullspaceeigmat}_0\T{\covnullspaceeigmat}_0) = {\covnullspaceeigmat}_0\T{\covnullspaceeigmat}_0 = \eye_{\dimconstraint \times \dimconstraint}$, ${\covnullspaceolmat}_1\T{\covnullspaceolmat}_1 = {\covnullspaceeigmat}_1\T({\covnullspaceeigmat}_0{\covnullspaceeigmat}_0\T{\covnullspaceeigmat}_1) = {\covnullspaceeigmat}_1\T{\covnullspaceeigmat}_1 = \eye_{\dimconstraint \times \dimconstraint}$, ... ${\covnullspaceolmat}_\infty\T{\covnullspaceolmat}_\infty = {\covnullspaceeigmat}_\infty\T({\covnullspaceeigmat}_0{\covnullspaceeigmat}_0\T{\covnullspaceeigmat}_\infty) = {\covnullspaceeigmat}_\infty\T {\covnullspaceeigmat}_\infty = \eye_{\dimconstraint \times \dimconstraint}$, and ${\covnullspaceolmat}_0 {\covnullspaceolmat}_0\T = {\covnullspaceeigmat}_0\T({\covnullspaceeigmat}_0{\covnullspaceeigmat}_0\T{\covnullspaceeigmat}_0) = {\covnullspaceeigmat}_0\T{\covnullspaceeigmat}_0 = \eye_{\dimconstraint \times \dimconstraint}$, ${\covnullspaceolmat}_1{\covnullspaceolmat}_1\T = {\covnullspaceeigmat}_0\T({\covnullspaceeigmat}_1{\covnullspaceeigmat}_1\T{\covnullspaceeigmat}_0) = {\covnullspaceeigmat}_0\T{\covnullspaceeigmat}_0 = \eye_{\dimconstraint \times \dimconstraint}$, ... ${\covnullspaceolmat}_\infty{\covnullspaceolmat}_\infty\T = {\covnullspaceeigmat}_0\T({\covnullspaceeigmat}_\infty{\covnullspaceeigmat}_\infty\T{\covnullspaceeigmat}_0) = {\covnullspaceeigmat}_0\T {\covnullspaceeigmat}_0 = \eye_{\dimconstraint \times \dimconstraint}$. 
All these show that ${\covnullspaceolmat}_0$, ${\covnullspaceolmat}_1$, ${\covnullspaceolmat}_2$, ..., ${\covnullspaceolmat}_\infty \in O(\dimconstraint)$.
Moreover, using ${\covnullspaceeigmat}_0$ as the inverse projection matrix, we get ${\covnullspaceeigmat}_0 = {\covnullspaceeigmat}_0 {\covnullspaceolmat}_0$, ${\covnullspaceeigmat}_1 = {\covnullspaceeigmat}_0 {\covnullspaceolmat}_1$, ${\covnullspaceeigmat}_2 = {\covnullspaceeigmat}_0 {\covnullspaceolmat}_2$, ... ${\covnullspaceeigmat}_\infty = {\covnullspaceeigmat}_0 {\covnullspaceolmat}_\infty$.
Therefore, there is a one-to-one mapping between ${\covnullspaceeigmat}_0$, ${\covnullspaceeigmat}_1$, ${\covnullspaceeigmat}_2$, ..., ${\covnullspaceeigmat}_\infty$ and ${\covnullspaceolmat}_0$, ${\covnullspaceolmat}_1$, ${\covnullspaceolmat}_2$, ..., ${\covnullspaceolmat}_\infty$.
Furthermore, between any two of 
${\covnullspaceeigmat}_0$, ${\covnullspaceeigmat}_1$, ${\covnullspaceeigmat}_2$, ..., ${\covnullspaceeigmat}_\infty$, e.g. ${\covnullspaceeigmat}_i$ and ${\covnullspaceeigmat}_j$, there exists $\diffson \in SO(\dimconstraint)$ such that ${\covnullspaceeigmat}_i \diffson = {\covnullspaceeigmat}_j$ if their $SO(\dimconstraint)$ projections ${\covnullspaceolmat}_i$ and ${\covnullspaceolmat}_j$ both are members of the same connected component of $O(\dimconstraint)$.

Now, suppose for nearby on-manifold data points $\jointposition_a$ and $\jointposition_c$, their approximate normal spaces $\normalspaceid_{\jointposition_a}\constraintmanifold$ and $\normalspaceid_{\jointposition_c}\constraintmanifold$ are spanned by eigenvector bases $\lnormalcoordframe^a = \{\coveigvec^a_{\dimambient-\dimconstraint+1}, \dots, \coveigvec^a_\dimambient\}$ and $\lnormalcoordframe^c = \{\coveigvec^c_{\dimambient-\dimconstraint+1}, \dots, \coveigvec^c_\dimambient\}$, respectively. Due to the curvature on the manifold $\constraintmanifold$, the normal spaces $\normalspaceid_{\jointposition_a}\constraintmanifold$ and $\normalspaceid_{\jointposition_c}\constraintmanifold$ may intersect, but in general are different subspaces of $\Re^{\dimambient \times \dimambient}$. 
For the purpose of aligning the basis of $\normalspaceid_{\jointposition_a}\constraintmanifold$ to the basis of $\normalspaceid_{\jointposition_c}\constraintmanifold$, one may think to do projection of the basis vectors of $\normalspaceid_{\jointposition_a}\constraintmanifold$ into $\normalspaceid_{\jointposition_c}\constraintmanifold$. Problematically, this projection may result in a non-orthogonal basis of $\normalspaceid_{\jointposition_c}\constraintmanifold$. 
Hence, we resort to an iterative method using a differentiable Special Orthogonal Group $SO(\dimconstraint)$. In particular, we form an $\dimconstraint \times \dimconstraint$ skew-symmetric matrix $\skewsymmmat \in so(\dimconstraint)$ with $\dimconstraint (\dimconstraint - 1) / 2$ differentiable parameters --where $so(\dimconstraint)$ is the Lie algebra of $SO(\dimconstraint)$, i.e. the set of all skew-symmetric $\dimconstraint \times \dimconstraint$ matrices--, 
and transform it through a differentiable exponential mapping (or matrix exponential) to get $\diffson = \exp(\skewsymmmat)$ with $\exp: so(\dimconstraint) \rightarrow SO(\dimconstraint)$. With $\covnullspaceeigmat^a = 
\begin{bmatrix}
    \coveigvec^a_{\dimambient-\dimconstraint+1} & \dots & \coveigvec^a_\dimambient
\end{bmatrix}$ and $\covnullspaceeigmat^c = 
\begin{bmatrix}
    \coveigvec^c_{\dimambient-\dimconstraint+1} & \dots & \coveigvec^c_\dimambient
\end{bmatrix}$, we can do an iterative training process to minimize the alignment error between $\covnullspaceeigmat^a \diffson$ and $\covnullspaceeigmat^c$, that is $\osaloss = \norm{\eye_{\dimconstraint \times \dimconstraint} - (\covnullspaceeigmat^a \diffson)\T \covnullspaceeigmat^c}_2^2$. Depending on whether both $\covnullspaceolmat^a$ and $\covnullspaceolmat^c$ (which are the projections of $\covnullspaceeigmat^a$ and $\covnullspaceeigmat^c$, respectively, to $O(\dimconstraint)$) are members of the same connected component of $O(\dimconstraint)$ or not, this alignment process may succeed or fail. 
However, if we define $\flippedcovnullspaceeigmat^a = 
\begin{bmatrix}
    -\coveigvec^a_{\dimambient-\dimconstraint+1} & \coveigvec^a_{\dimambient-\dimconstraint+2} & \dots & \coveigvec^a_\dimambient
\end{bmatrix}$ and $\flippedcovnullspaceeigmat^c = 
\begin{bmatrix}
    -\coveigvec^c_{\dimambient-\dimconstraint+1} & \coveigvec^c_{\dimambient-\dimconstraint+2} & \dots & \coveigvec^c_\dimambient
\end{bmatrix}$, two out of the four pairs $(\covnullspaceeigmat^a, \covnullspaceeigmat^c)$, $(\flippedcovnullspaceeigmat^a, \covnullspaceeigmat^c)$, $(\covnullspaceeigmat^a, \flippedcovnullspaceeigmat^c)$, and $(\flippedcovnullspaceeigmat^a, \flippedcovnullspaceeigmat^c)$ will be pairs in the same connected component. Thus, two of these pairs will achieve minimum alignment errors after training the differentiable Special Orthogonal Groups $SO(\dimconstraint)$ on these pairs, indicating successful alignment. These are the main insights for our \emph{local} alignment of neighboring normal spaces of on-manifold data points.

For the \emph{global} alignment of the normal spaces, we represent the on-manifold data points as a graph. 
Our Orthogonal Subspace Alignment (OSA) is outlined in Algorithm~\ref{alg:osa}. 
We begin by constructing a sparse graph of nearest neighbor connections of each on-manifold data point, followed by the construction of this graph into an (un-directed) minimum spanning tree (MST), and eventually the conversion of the MST to a directed acyclic graph (DAG). This graph construction is detailed in lines~\ref{algl:startgraphconstruction} - \ref{algl:endgraphconstruction} of Algorithm~\ref{alg:osa}.

Each directed edge in the DAG represents a pair of on-manifold data points whose normal spaces are to be aligned locally.
Our insights for the local alignment of neighboring normal spaces are implemented in lines~\ref{algl:startlocalalignment} - \ref{algl:endlocalalignment} of Algorithm~\ref{alg:osa}.
In the actual implementation, these local alignment computations are done as a vectorized computation which is faster than doing it in a for-loop as presented in Algorithm~\ref{alg:osa}; this for-loop presentation is made only for the sake of clarity. 
We initialize the $\dimconstraint (\dimconstraint - 1) / 2$ differentiable parameters of the $\dimconstraint \times \dimconstraint$ skew-symmetric matrix $\skewsymmmat$ with near zero random numbers, which essentially will map to a near identity matrix $\eye_{\dimconstraint \times \dimconstraint}$ of $\diffson$ via the $\exp()$ mapping, as stated in line~\ref{algl:diffsoninit} of Algorithm~\ref{alg:osa}\footnote{Although most of our {\ecmnn} implementation is done in PyTorch \cite{Paszke_PyTorch_AutoDiff_2017}, the OSA algorithm is implemented in TensorFlow \cite{TensorFlowBib}, because at the time of implementation of the OSA algorithm, PyTorch did not support the differentiable matrix exponential (i.e. the exponential mapping) computation yet while TensorFlow did.}. This is reasonable because we assume that most of the neighboring normal spaces are already/close to being aligned initially. We optimize the alignment of the four pairs $(\covnullspaceeigmat^a, \covnullspaceeigmat^c)$, $(\flippedcovnullspaceeigmat^a, \covnullspaceeigmat^c)$, $(\covnullspaceeigmat^a, \flippedcovnullspaceeigmat^c)$, and $(\flippedcovnullspaceeigmat^a, \flippedcovnullspaceeigmat^c)$ in lines~\ref{algl:firstiterativealignmenterrorminimization} - \ref{algl:fourthiterativealignmenterrorminimization} of Algorithm~\ref{alg:osa}. 

Once the local alignments are done, the algorithm then traverses the DAG in breadth-first order, starting from the root $\rootid$, where the orientation of the root is already chosen and committed to. During the breadth-first traversal of the DAG, three things are done: First, the orientation of each point is chosen based on the minimum alignment loss; second, the local alignment transforms are compounded/integrated along the path from root to the point; and finally, the (globally) aligned orthogonal basis of each point is computed and returned as the result of the algorithm. These steps are represented by lines~\ref{algl:startglobalalignment} -- \ref{algl:endglobalalignment} of Algorithm~\ref{alg:osa}.

\begin{algorithm}[H]
	\caption{Orthogonal Subspace Alignment (OSA)}
	\label{alg:osa}
	\begin{algorithmic}[1]
	    \Function{OSA}{$\{(\jointposition \in \onconstraintconfigspace, \text{orthogonal basis stacked as matrix } \covnullspaceeigmat \text{ associated with } \normalspaceatjointposition)\}$}
    		\State \label{algl:startgraphconstruction} \# construct a sparse graph between each data point $\jointposition \in \onconstraintconfigspace$ with its $\mstnumnearestneighbor$ nearest neighbors, 
    		\State \# followed by minimum spanning tree and directed acyclic graph computations 
    		\State \# to obtain directed edges $\dagedges$; $\mstnumnearestneighbor$ needs to be chosen to be a value as small as possible that 
    		\State \# still results in all non-root points $\{\jointposition \in \onconstraintconfigspace \backslash \{\rootjointposition\} \}$ being reachable from the root point $\rootjointposition$:
    		\State $\nnsparsegraph \gets \textit{computeNearestNeighborsSparseGraph}(\{\jointposition \in \onconstraintconfigspace\}, \mstnumnearestneighbor)$
    		\State $\mst \gets \textit{computeMinimumSpanningTree}(\nnsparsegraph)$
    		\State \label{algl:endgraphconstruction} $\dagedges \gets \textit{computeDirectedAcyclicGraphEdgesByBreadthFirstTree}(\mst)$
    		\For{ each directed edge $\directededge = (\jointposition_c, \jointposition_a) \in \dagedges$} \label{algl:startlocalalignment}
    		    \State Obtain $\covnullspaceeigmat^a = 
                \begin{bmatrix}
                    \coveigvec^a_{\dimambient-\dimconstraint+1} & \dots & \coveigvec^a_\dimambient
                \end{bmatrix} \in \Re^{\dimambient \times \dimconstraint}$ associated with the source subspace $\normalspaceid_{\jointposition_a}\constraintmanifold$ 
    		    \State Obtain $\covnullspaceeigmat^c = 
                \begin{bmatrix}
                    \coveigvec^c_{\dimambient-\dimconstraint+1} & \dots & \coveigvec^c_\dimambient
                \end{bmatrix} \in \Re^{\dimambient \times \dimconstraint}$ associated with the target subspace $\normalspaceid_{\jointposition_c}\constraintmanifold$
                \State Define $\flippedcovnullspaceeigmat^a = 
                \begin{bmatrix}
                    -\coveigvec^a_{\dimambient-\dimconstraint+1} & \coveigvec^a_{\dimambient-\dimconstraint+2} & \dots & \coveigvec^a_\dimambient
                \end{bmatrix} \in \Re^{\dimambient \times \dimconstraint}$
                \State Define $\flippedcovnullspaceeigmat^c = 
                \begin{bmatrix}
                    -\coveigvec^c_{\dimambient-\dimconstraint+1} & \coveigvec^c_{\dimambient-\dimconstraint+2} & \dots & \coveigvec^c_\dimambient
                \end{bmatrix} \in \Re^{\dimambient \times \dimconstraint}$
                \State \label{algl:diffsoninit} Define differentiable $SO(\dimconstraint)$ $\diffson^{\overrightarrow{a}\overrightarrow{c}}$, $\diffson^{\overrightarrow{a}\overleftarrow{c}}$, $\diffson^{\overleftarrow{a}\overrightarrow{c}}$, and $\diffson^{\overleftarrow{a}\overleftarrow{c}}$, initialized near identity
                \State \# try optimizing the alignment of the 4 possible pairs:
                \State \label{algl:firstiterativealignmenterrorminimization} $(\diffson^{\overrightarrow{a}\overrightarrow{c}}, \loss_{\overrightarrow{a}\overrightarrow{c}}) \gets \textit{iterativelyMinimizeAlignmentError}(\covnullspaceeigmat^a \diffson^{\overrightarrow{a}\overrightarrow{c}}, \covnullspaceeigmat^c)$
                \State $(\diffson^{\overrightarrow{a}\overleftarrow{c}}, \loss_{\overrightarrow{a}\overleftarrow{c}}) \gets \textit{iterativelyMinimizeAlignmentError}(\covnullspaceeigmat^a \diffson^{\overrightarrow{a}\overleftarrow{c}}, \flippedcovnullspaceeigmat^{c})$
                \State $(\diffson^{\overleftarrow{a}\overrightarrow{c}}, \loss_{\overleftarrow{a}\overrightarrow{c}}) \gets \textit{iterativelyMinimizeAlignmentError}(\flippedcovnullspaceeigmat^{a} \diffson^{\overleftarrow{a}\overrightarrow{c}}, \covnullspaceeigmat^c)$
                \State \label{algl:fourthiterativealignmenterrorminimization} $(\diffson^{\overleftarrow{a}\overleftarrow{c}}, \loss_{\overleftarrow{a}\overleftarrow{c}}) \gets \textit{iterativelyMinimizeAlignmentError}(\flippedcovnullspaceeigmat^{a} \diffson^{\overleftarrow{a}\overleftarrow{c}}, \flippedcovnullspaceeigmat^{c})$
                \State \# record optimized local alignment rotation matrices and its associated loss w/ the edge: 
                \State \label{algl:endlocalalignment} Associate $(\diffson^{\overrightarrow{a}\overrightarrow{c}}, \loss_{\overrightarrow{a}\overrightarrow{c}})$, $(\diffson^{\overrightarrow{a}\overleftarrow{c}}, \loss_{\overrightarrow{a}\overleftarrow{c}})$, $(\diffson^{\overleftarrow{a}\overrightarrow{c}}, \loss_{\overleftarrow{a}\overrightarrow{c}})$, $(\diffson^{\overleftarrow{a}\overleftarrow{c}}, \loss_{\overleftarrow{a}\overleftarrow{c}})$ with $\directededge$
    		\EndFor
    		\State \label{algl:startglobalalignment} \# commit on the orientation of the root point as un-flipped ($\overrightarrow{\rootid}$) instead of flipped ($\overleftarrow{\rootid}$):
    		\State $ori(\rootid) = \overrightarrow{\rootid}$
    		\State \# define the compound/global alignment rotation matrix of the root as an identity matrix:
    		\State $\globalalignmentrotmat^{\rootid} = \eye_{\dimconstraint \times \dimconstraint}$
    		\State \# aligned orthogonal basis of the root is:
    		\State ${\covnullspaceeigmat}^{\text{aligned}, \rootid} = \covnullspaceeigmat^{\rootid}$
    		\State \# do breadth-first traversal from root to:
    		\State \# (1) select the orientation $ori()$ of each point based on the minimum alignment loss, 
    		\State \# (2) compound/integrate the local alignment transforms $\globalalignmentrotmat$ along the path to the point,  
    		\State \# (3) and finally compute the aligned orthogonal basis $\alignedcovnullspaceeigmat$:
    		\State $Q = \textit{Queue}()$
    		\State $Q.\textit{enqueue}(\textit{childrenOfNodeInGraph}(\rootid, \dagedges))$
    		\While{$\textit{size}(Q) > 0$}
    		    \State $\currentnodeid = Q.\textit{dequeue}()$
    		    \State $Q.\textit{enqueue}(\textit{childrenOfNodeInGraph}(\currentnodeid, \dagedges))$
    		    \State $\parentnodeid = \textit{parentOfNodeInGraph}(\currentnodeid, \dagedges)$
    		    \State \# select the local alignment rotation matrix based on the minimum alignment loss 
    		    \State \# among the two possibilities:
    		    \If{$\loss_{\overrightarrow{\currentnodeid}ori(\parentnodeid)} < \loss_{\overleftarrow{\currentnodeid}ori(\parentnodeid)}$}
    		        \State $ori(\currentnodeid) = \overrightarrow{\currentnodeid}$
    		        \State $\globalalignmentrotmat^{\currentnodeid} = \diffson^{\overrightarrow{\currentnodeid}ori(\parentnodeid)} \globalalignmentrotmat^{\parentnodeid}$
    		        \State ${\covnullspaceeigmat}^{\text{aligned}, \currentnodeid} = \covnullspaceeigmat^{\currentnodeid} \globalalignmentrotmat^{\currentnodeid}$
    		    \Else
    		        \State $ori(\currentnodeid) = \overleftarrow{\currentnodeid}$
    		        \State $\globalalignmentrotmat^{\currentnodeid} = \diffson^{\overleftarrow{\currentnodeid}ori(\parentnodeid)} \globalalignmentrotmat^{\parentnodeid}$
    		        \State ${\covnullspaceeigmat}^{\text{aligned}, \currentnodeid} = \flippedcovnullspaceeigmat^{\currentnodeid} \globalalignmentrotmat^{\currentnodeid}$
    		    \EndIf
    		\EndWhile
    		\State \label{algl:endglobalalignment} \Return $\{\alignedcovnullspaceeigmat \text{ associated with } \normalspaceatjointposition \text{ for each } \jointposition \in \onconstraintconfigspace\}$
    	\EndFunction
	\end{algorithmic}
\end{algorithm}

\newpage
\subsection{Additional Experiments}
\label{sec:addtnl_exp}
\new{\subsubsection{Learning {\ecmnn} on noisy data}
We also evaluate {\ecmnn} learning on noisy data. We generate a noisy unit sphere dataset and a noisy 3D unit circle with additive Gaussian noise of zero mean and standard deviation 0.01. After we train {\ecmnn} on these noisy sphere and 3D circle datasets, we evaluate the model and obtain (82.00 $\pm$ 12.83) \% and (89.33 $\pm$ 11.61) \%, respectively, as the $P_{\bar{h}_M}$ metric.\footnote{\new{The small positive scalar $\augmagnitude$ needs to be chosen sufficiently large as compared to the noise level, so that the data augmentation will not create inconsistent data w.r.t. the noise.}}
These are still quite high success rates and outperform the VAE without noise.}
\new{\subsubsection{Relationship between the number of augmentation levels and the projection success rate}
We also perform an experiment to study the relationship between the number of augmentation levels (the maximum value of the positive integer $\augidx$ in the off-manifold points $\augjointposition = \jointposition + \augidx \augmagnitude \randunitvec$) and the projection success rate. As we vary this parameter at 1, 2, 3, and 7 on the Sphere dataset, the projection success rates are (5.00 $\pm$ 2.83) \%, (12.33 $\pm$ 9.03) \%, (83.67 $\pm$ 19.01) \%, and (97.33 $\pm$ 3.77) \%, respectively, showing that the projection success rate improves as the number of augmentation levels are increased.
Increasing this parameter too high, however, would eventually have two problems: First, we empirically found data augmentation to be a computationally expensive step in training, and second, it would be possible to run into an issue like augmenting a point on a sphere beyond the center of a sphere (as mentioned in section \ref{ssec:data_augmentation}).}

\end{document}